\documentclass{article} %
\usepackage[dvipsnames]{xcolor}
\usepackage{iclr2021_conference,times}

\usepackage{amsmath,amsfonts,bm}

\def\eqref#1{equation~\ref{#1}}

\def\1{\bm{1}}

\DeclareMathAlphabet{\mathsfit}{\encodingdefault}{\sfdefault}{m}{sl}
\SetMathAlphabet{\mathsfit}{bold}{\encodingdefault}{\sfdefault}{bx}{n}

\newcommand{\E}{\mathbb{E}}

\newcommand{\R}{\mathbb{R}}

\usepackage{hyperref}
\usepackage{url}
\usepackage{wrapfig}
\usepackage{calc}
\newcounter{fignum}

\usepackage{amsmath}
\usepackage{bm}
\usepackage{graphicx}
\usepackage{booktabs}
\usepackage{xcolor}
\usepackage{caption}
\usepackage{subcaption}
\usepackage{tikz-network} 

\newcommand{\meth}[1]{\textsc{\footnotesize{#1}}}

\newcommand*{\method}{\meth{Amorpheus}}
\newcommand{\Method}{\textsc{Amorpheus}}

\newcommand{\calA}{\mathcal{A}}

\newcommand{\calS}{\mathcal{S}}
\newcommand{\calV}{\mathcal{V}}
\newcommand{\calR}{\mathcal{R}}
\newcommand{\calT}{\mathcal{T}}

\newcommand{\Set}[1]{\mathcal{#1}}
\renewcommand{\vec}[1]{\bm{#1}}
\def\R{\mathop{\mathbb{R}}}

\usepackage[acronym,smallcaps,nowarn,section,nogroupskip,nonumberlist, automake]{glossaries}    
\setkeys{glslink}{hyper=false}    
\makeglossaries
\newacronym{rl}{\meth{RL}}{Reinforcement Learning}    
\newacronym{gnn}{\meth{GNN}}{Graph Neural Network} 
\newacronym{mdp}{\meth{MDP}}{Markov Decision Process} 
\newacronym{pomdp}{\meth{POMDP}}{Partially Observable Markov Decision Process} 
\newacronym{mtrl}{\meth{MTRL}}{Multitask Reinforcement Learning}
\newacronym{drl}{\meth{DRL}}{Deep Reinforcement Learning}
\newacronym{smp}{\meth{SMP}}{Shared Modular Policies}
\newacronym{gru}{\meth{GRU}}{Gated Recurrent Unit}
\newacronym{mlp}{\meth{MLP}}{Multi-layer Perceptron}
\newacronym{pg}{\meth{PG}}{Policy Gradient}

\usepackage{soul}

\definecolor{darkgreen}{rgb}{0, 0.75, 0}

\usepackage[colorinlistoftodos, textsize=tiny, textwidth=34mm]{todonotes}
\definecolor{wbcolor}{rgb}{0.5, 1, 0.5}
\newcommand{\wbs}[1]{}

\newcommand{\wbr}[2]{#1}

\title{My Body is a Cage: the Role of Morphology\\in Graph-Based Incompatible Control}

\author{Vitaly Kurin \\
Department of Computer Science\\
University of Oxford\\
Oxford, United Kingdom \\
\texttt{vitaly.kurin@cs.ox.ac.uk} \\
\And
Maximilian Igl \\
Department of Computer Science\\
University of Oxford\\
Oxford, United Kingdom \\
\texttt{maximilian.igl@eng.ox.ac.uk} \\
\And
Tim Rockt\"aschel\\
Department of Computer Science\\
University College London\\
London, United Kingdom \\
\texttt{t.rocktaschel@cs.ucl.ac.uk} \\
\And
Wendelin B\"ohmer \\
Department of Software Technology\\
Delft University of Technology\\
Delft, Netherlands\\
\texttt{j.w.bohmer@tudelft.nl} \\
\And
Shimon Whiteson \\
Department of Computer Science\\
University of Oxford\\
Oxford, United Kingdom \\
\texttt{shimon.whiteson@cs.ox.ac.uk} \\
}

\iclrfinalcopy %
\begin{document}

\maketitle
\begin{abstract}
Multitask Reinforcement Learning is a promising way to obtain models with better performance, generalisation, data efficiency, and robustness.
Most existing work is limited to \emph{compatible} settings, where the state and action space dimensions are the same across tasks.
Graph Neural Networks (GNN) are one way to address incompatible environments, because they can process graphs of arbitrary size.
They also allow practitioners to inject biases encoded in the structure of the input graph.
Existing work in graph-based continuous control uses the physical morphology of the agent to construct the input graph, i.e., encoding limb features as node labels and using edges to connect the nodes if their corresponded limbs are physically connected.
In this work, we present a series of ablations on existing methods that show that 
morphological information encoded in the graph does not improve their
performance.
Motivated by the hypothesis that any benefits GNNs extract from the graph structure are outweighed by difficulties they create for message passing, we also propose \method{}, a transformer-based approach. Further results show that, while \method{} ignores the morphological information that GNNs encode, it nonetheless substantially outperforms GNN-based methods that use the morphological information to define the message-passing scheme.
\end{abstract}

\section{Introduction}

\gls{mtrl} \citep{varghese2020mtrl} leverages commonalities between multiple tasks to obtain policies with better returns, generalisation, data efficiency, or robustness.
Most \gls{mtrl} work assumes {\em compatible} state-action spaces,
where the dimensionality of the states and actions is the same across tasks.
However, many practically important domains, such as robotics, combinatorial optimization,
and object-oriented environments,
have {\em incompatible} state-action spaces 
and cannot be solved by common \gls{mtrl} approaches.

Incompatible environments are avoided largely because they are inconvenient for function approximation: conventional architectures expect fixed-size inputs and outputs.
One way to overcome this limitation is to use \glspl{gnn}~\citep{gori2005new, scarselli2005graph, battaglia2018relational}.
A key feature of \glspl{gnn} is that they can process graphs of arbitrary size and thus, in principle, allow~\gls{mtrl} in incompatible environments.
However, \Glspl{gnn} also have a second key feature: they allow models to condition on structural information 
about how state features are related, e.g., how a robot's limbs are connected.
In effect, this enables practitioners to incorporate additional domain knowledge where states are described as labelled graphs. 
Here, a graph is a collection of labelled nodes, indicating the features of corresponding objects, and edges, indicating the relations between them.
In many cases, e.g., with  the robot mentioned above, such domain knowledge is readily available.
This results in a structural inductive bias that restricts the model's computation graph, determining how errors backpropagate through the network.

\glspl{gnn} have been applied to \gls{mtrl} in continuous control environments, a staple benchmark of modern~\gls{rl}, by leveraging both of the key features mentioned above~\citep{wang2018nervenet, huang2020smp}.  
In these two works, the labelled graphs are based on the agent's physical morphology, with nodes labelled with the observable features of their corresponding limbs, e.g., coordinates, angular velocities and limb type.
If two limbs are physically connected, there is an edge between their corresponding nodes.
However, the assumption that it is beneficial to restrict the model's
computation graph in this way has to our knowledge not been validated.

To investigate this issue, we conduct a series of ablations on existing \gls{gnn}-based continuous control methods. The results show that removing morphological information does not harm the performance of these models.
In addition, we propose \method{}, a new continuous control MTRL method based on transformers~\citep{vaswani2017attention} instead of \glspl{gnn} that use morphological information to define the message-passing scheme. 
\method{}~is motivated by the hypothesis that any benefit \glspl{gnn} can extract from the morphological domain knowledge encoded in the graph is outweighed by the difficulty that the graph creates for message passing.  
In a sparsely connected graph, crucial state information must be communicated across multiple hops, which we hypothesise is difficult in practice to learn.  
\method~uses transformers instead, which can be thought of as fully connected \glspl{gnn} with attentional aggregation~\citep{battaglia2018relational}.  
Hence, \method{}~ignores the morphological domain knowledge but in exchange obviates the need to learn multi-hop communication.
Similarly, in Natural Language Processing, transformers were shown to perform better without an explicit structural bias and even learn such structures from data~\citep{vig2019analyzing, goldberg2019assessing, tenney2019you, peters2018dissecting}.

Our results on incompatible \gls{mtrl}~continious control benchmarks \citep{huang2020smp,wang2018nervenet} strongly support our hypothesis: \method~substantially outperforms \gls{gnn}-based alternatives with fixed message-passing schemes in terms of sample efficiency and final performance.  
In addition, \method~exhibits nontrivial behaviour such as cyclic attention patterns coordinated with gaits.
\section{Background}

We now describe the necessary background for the rest of the paper.

\subsection{Reinforcement Learning}
\label{sec:mdp-formalism}

A~\gls{mdp} is a tuple $\langle \calS, \calA, \calR, \calT, \rho_0 \rangle$.
The first two elements define the set of states $\calS$ and the set of actions $\calA$.
The next element defines the reward function $\calR(s,a,s')$ with $s, s'\in \calS$ and $a \in \calA$.
$\calT(s'|s,a)$ is the probability distribution function over states $s' \in \calS$ after taking action $a$ in state $s$. 
The last element of the tuple $\rho_0$ is the distribution over initial states.
Task and environment are synonyms for \glspl{mdp} in this work.

A policy $\pi(a|s)$ is a mapping from states to distributions over actions.
The goal of an \meth{RL} agent is to find a policy that maximises the expected discounted cumulative return $J = \E\big[\sum_{t=0}^{\infty} \gamma^t r_t\big]$, where $\gamma \in [0,1)$ is a discount factor, $t$ is the discrete environment step and $r_t$ is the reward at step $t$. 
In the \gls{mtrl} setting, the agent aims to maximise the average performance across $N$ tasks: $\frac{1}{N}\sum_{i=1}^N{J_i}$.
We use \emph{\gls{mtrl} return} to denote the average performance across the tasks.

In this paper, we assume that \wbr{states and actions are multivariate, but dimensionality remains constant for one \gls{mdp}}{the state and action sets elements of an \gls{mdp} are of constant dimension}: $s \in \R^k,\forall s \in \calS$, and $a \in \R^{k'},\forall a \in \calA$.
We use $dim(\calS)=k$ and $dim(\calA)=k'$ to denote this dimensionality, which can differ amongst \glspl{mdp}.
We consider two tasks \gls{mdp}$_1$  and \gls{mdp}$_2$ as \emph{incompatible} if the dimensionality of their state or action spaces disagree, i.e., 
$dim(\calS_1)\neq dim(\calS_2)$ or $dim(\calA_1)\neq dim(\calA_2)$ with the subscript denoting a task index.
In this case \gls{mtrl} policies or value functions
can not be represented by a \gls{mlp},
which requires fixed input dimensions. 
We do not have additional assumptions on the semantics behind the state and action set elements and focus on the dimensions mismatch only.

Our approach, as well as the baselines in this work~\citep{wang2018nervenet,huang2020smp}, use~\gls{pg} methods~\citep{peters2006policy}.
\gls{pg} methods optimise a policy using gradient ascent on the objective: $\theta_{t+1} = \theta_t + \alpha \nabla_{\theta}J |_{\theta=\theta_t}$, where $\theta$ parameterises a policy.
Often, to reduce variance in the gradient estimates, one learns a critic so that the policy gradient becomes $\nabla_{\theta}J(\theta) \!=\! \E \big[\sum_{t} A^{\pi}_t \, \nabla_{\theta} \log\pi_{\theta}(a_t|s_t) \big]$, where $A^{\pi}_t$ is an estimate of the advantage function (e.g., TD residual $r_t + \gamma V^{\pi}(s_{t+1}) - V^{\pi}(s_{t})$).
The state-value function $V^{\pi}(s)$ is the expected discounted return a policy $\pi$ receives starting at state $s$.
\citet{wang2018nervenet} use PPO~\citep{schulman2017proximal}, which restricts a policy update to avoid instabilities from drastic changes in the policy behaviour.
\citet{huang2020smp} use \meth{TD3}~\citep{fujimoto2018addressing}, a~\gls{pg} method based on
\meth{DDPG}~\citep{lillicrap2015continuous}.

\subsection{Graph Neural Networks for Incompatible Multitask RL}
\def\concat{\bar}                       
\glspl{gnn} can address incompatible environments because they can process graphs of arbitrary sizes and topologies.
A~\gls{gnn} is a function that takes a labelled graph as input
and outputs a graph $\Set G'$ with different labels but the same topology.
Here, a labelled graph $\Set G := \langle \Set V, \Set E \rangle$
consists of a set of vertices $v^i \in \Set V$, 
labelled with vectors $\vec v^i \in \R^{m_v}$ and
a set of directed edges $e^{ij} \in \Set E$ 
from vertex $v^i$ to $v^j$,
labelled with vectors $\vec e^{ij} \in \R^{m_e}$.
The output graph $\Set G'$
has the same topology but the labels can 
be of different dimensionality than the input,
that is, $\vec v'^i \in \R^{m'_v}$ and 
$\vec e'^{ij} \in \R^{m'_e}$.
By graph topology we mean the connectivity of the graph, which can be represented by an adjacency matrix, a binary matrix $\{a\}_{ij}$ whose elements $a_{ij}$ equal to one iff there is an edge $e_{ij} \in \Set E$ connecting vertices $v_i, v_j \in \Set V$.

A \gls{gnn} computes the output labels for entities of type $k$
by parameterised {\em update functions} $\phi_\psi^k$ represented by neural networks that can be learnt end-to-end via backpropagation.
These updates can depend on a varying number of edges or vertices,
which have to be summarised first using {\em aggregation functions} that we denote $\rho$.
Apart from their ability to operate on sets of elements, aggregation functions should be permutation invariant.
Examples of such aggregation functions include summation, averaging and $\max$ or $\min$ operations.

{Incompatible} \gls{mtrl} for continuous control implies learning a common policy for a set of agents with different number of limbs and connectivity of those limbs, i.e. \emph{morphology}.
To be more precise, a set of incompatible continuous control environments is a set of \glspl{mdp} described in Section~\ref{sec:mdp-formalism}.
When a state is represented as a graph, each node label contains features of its corresponding limb, e.g., limb type, coordinates, and angular velocity.
Similarly, each factor of an action set element corresponds to a node with the label meaning the torque for a joint to emit.
The typical reward function of a MuJoCo~\citep{todorov2012mujoco} environment includes a reward for staying alive, distance covered, and a penalty for action magnitudes.

We now describe two existing approaches to incompatible control: 
\meth{NerveNet}~\citep{wang2018nervenet} and \gls{smp}~\citep{huang2020smp}.

\subsubsection{NerveNet}

In \meth{NerveNet}, the input observations are first encoded via a \gls{mlp} processing each node labels as a batch element: 
$\vec v^{i} \leftarrow \phi_\chi\big(\, {\vec v}^i\big), \forall v^{i} \in \Set V$.
After that, the message-passing part of the model block performs the following computations (in order):
\def\gnngap{\\[1mm]}
\begin{equation*} \label{eq:nervenet_update}
\begin{array}{rcl}
	\vec e'^{ij} 
	&\leftarrow& 
	\phi^e_\psi\big(\, {\vec v}^i\big) 
	\hfill, \forall e^{ij} \in \Set E \,,\\
	\vec v^i \;
	&\leftarrow& 
	\phi^v_\xi\big(\, {\vec v}^i, \,
		\rho\{\vec e'^{ki} \,|\, e^{ki} \in \Set E\}
	\big) 
	\quad, \forall v^i \in \Set V \,.
\end{array}
\end{equation*}
The edge updater $\phi^e_{\psi}$ in \meth{NerveNet} is an \gls{mlp} which does not take the receiver's state into account.
Using only one message pass restricts the learned function to local computations on the graph.
The node updater $\phi^v_{\xi}$ is a \gls{gru}~\citep{cho2014learning} which maintains the internal state when doing multiple message-passing iterations, and takes the aggregated outputs of the edge updater for all incoming edges as inputs.
After the message-passing stage, the \gls{mlp} decoder takes the states of the nodes and, like the encoder, independently processes them, emitting scalars used as the mean for the normal distribution from which actions are sampled: $\vec v_{dec}^{i} \leftarrow \phi_\eta\big(\, {\vec v}^i\big), \forall v^{i} \in \Set V$.
The standard deviation of this distribution is a separate state-independent vector with one scalar per action.

\subsubsection{Shared Modular Policies}

\gls{smp} is a variant of a \gls{gnn} that operates only on trees. Computation is performed in two stages: top-down and bottom-up.
In the first stage, information propagates level by level from leaves to the root with parents aggregating information from their children.
In the second stage, information propagates from parents to the leaves with parents emitting multiple messages, one per child.
The policy emits actions at the second stage of the computation together with the downstream messages.

Instead of a permutation invariant aggregation, the messages are concatenated.
This, as well as separate messages for the children, also injects structural bias to the model, e.g., separating the messages for the left and right parts of robots with bilateral symmetry.
In addition, its message-passing schema depends on the morphology and the choice of the root node.
In fact, \citet{huang2020smp} show that the root node choice can affect performance by 15\%.

\gls{smp} trains a separate model for the actor and critic.
An actor outputs one action per non-root node.
The critic outputs a scalar per node as well.
When updating a critic, a value loss is computed independently per each node with targets using the same scalar reward from the environment.

\subsection{Transformers}

Transformers can be seen as \glspl{gnn} applied to fully connected graphs  with the attention as an edge-to-vertex aggregation operation~\citep{battaglia2018relational}.
Self-attention used in transformers is an associative memory-like mechanism that first projects the feature vector of each node $\vec v^i \in \R^{m_v}$ into three vectors: query $\vec q_i := \vec \Theta \vec v^{i} \in \R^{d}$, key $\vec k_i := \bar{\vec \Theta} \vec v^{i} \in \R^d$ and value $\hat{\vec v}_i := \hat{\vec \Theta} \vec v^i \in \R^{m_v}$.
Parameter matrices $\vec \Theta, \bar{\vec \Theta} \text{, and } \hat{\vec
  \Theta}$ are learnt.
The query of the receiver $v_i$ is compared to the key value of senders using a dot product.
The resulting values $\vec w_i$ are used as weights in the weighted sum of all the value vectors in the graph.
The computation proceeds as follows:
\begin{equation}
	\begin{array}{rcl}
	\vec w_i &:=& \text{softmax}\big(\frac{[\vec k_1, \ldots, \vec k_n]^\top \vec q_i}{\sqrt{d}}\big)  \\
	\vec v'_i &:=& [\hat{\vec v}_1, \ldots, \hat{\vec v}_n] \vec w_{i} 
	\end{array}, \forall v_i \in \calV \,,
	\label{eq:attention}
\end{equation}
with $[x_1, x_2, ..., x_n]$ being a $\R^{k \times n}$ matrix of concatenated vectors $x_i \in \R^k$.
Often, multiple attention heads, i.e., $\vec \Theta, \bar{\vec \Theta} \text{, and } \hat{\vec \Theta}$ matrices, are used to learn different interactions between the nodes and mitigate the consequences of unlucky initialisation.
The output of multiple heads is concatenated and later projected to respect the dimensions.

A transformer block is a combination of an attention block and a feedforward layer with a possible normalisation between them.
In addition, there are residual connections from the input to the attention output and from the output of the attention to the feedforward layer output.
Transformer blocks can be stacked together to take higher order dependencies into account, i.e., reacting not only to the features of the nodes, but how the features of the nodes change after applying a transformer block.
\section{The Role of Morphology in Existing Work}
\label{sec:contribution}

In this section, we provide evidence against the assumption that \glspl{gnn} improve performance by exploiting information about physical morphology~\citep{huang2020smp,wang2018nervenet}.
Here and in all of the following sections, we run experiments for three random seeds and report the average undiscounted \gls{mtrl}~return and the standard error across the seeds.
\begin{figure}[h] %
    \centering
    \begin{subfigure}[t]{0.32\textwidth}
        \centering
        \includegraphics[width=\textwidth]{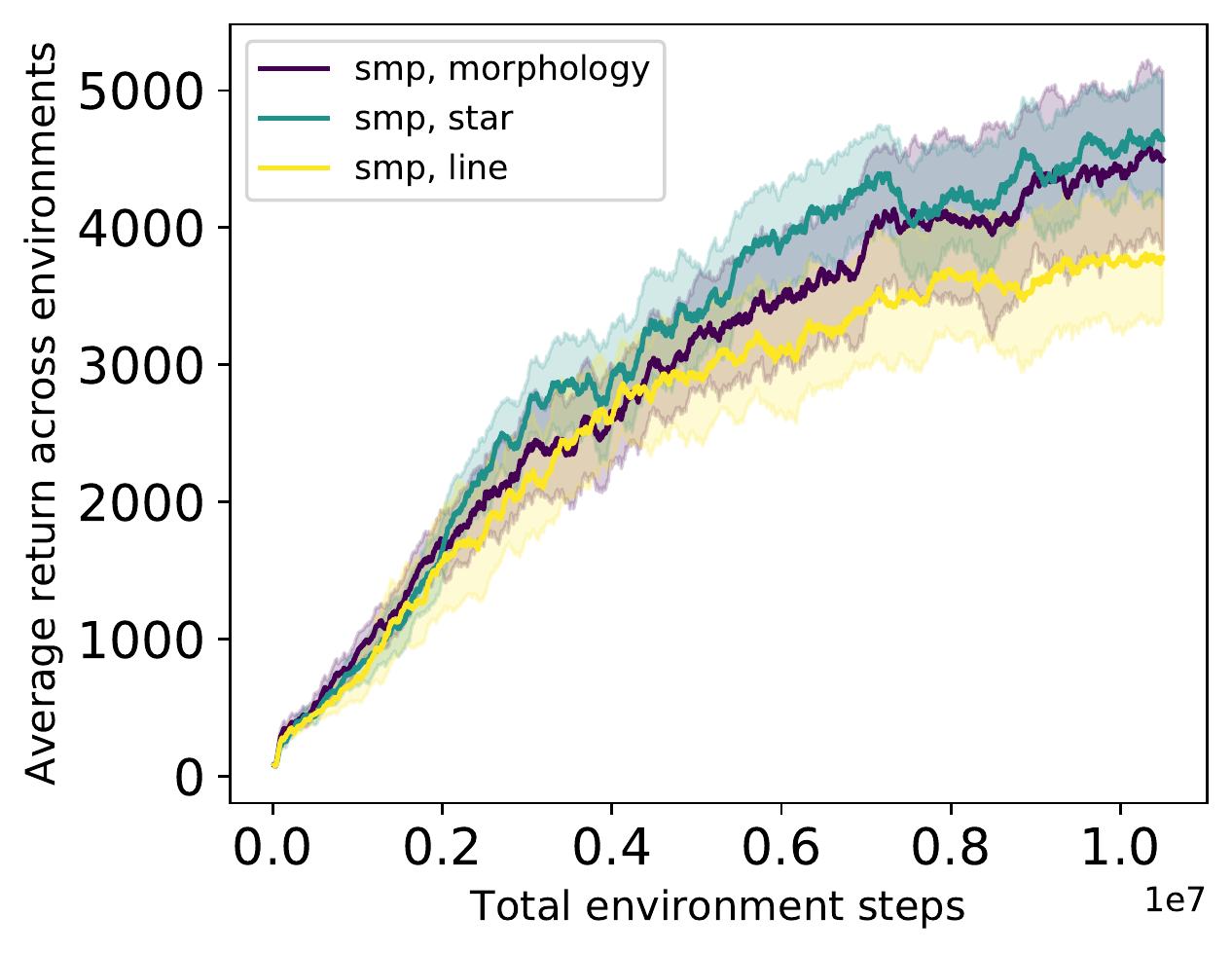}
        \caption{\gls{smp}, \texttt{Walker++}}
        \label{fig:structure-ablations-a}
    \end{subfigure}
    \hfill
    \begin{subfigure}[t]{0.32\textwidth}
        \centering
        \includegraphics[width=\textwidth]{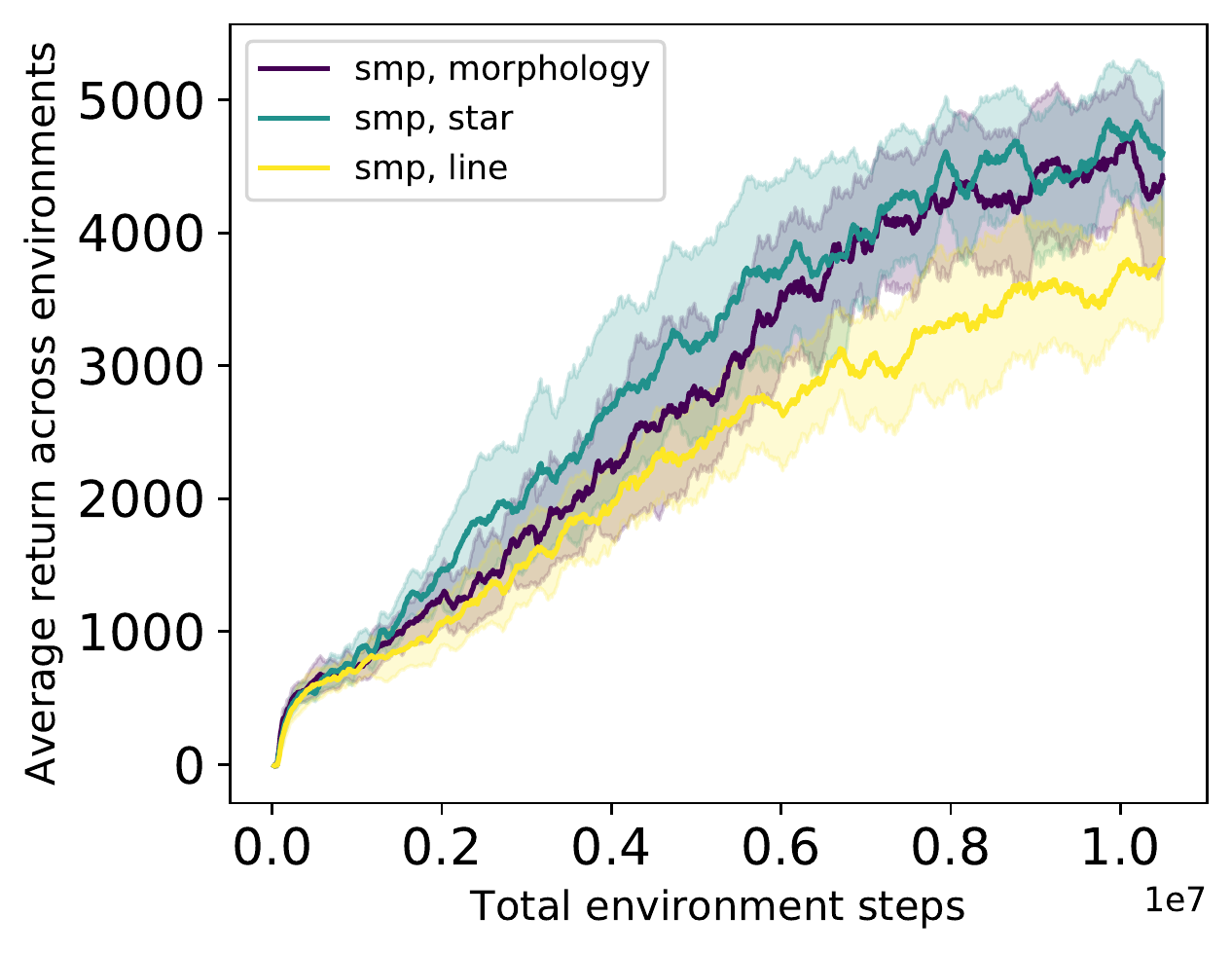}
        \caption{\gls{smp}, \texttt{Humanoid++}}
        \label{fig:structure-ablations-b}
    \end{subfigure}
    \hfill
    \begin{subfigure}[t]{0.32\textwidth}
        \centering
        \includegraphics[width=\textwidth]{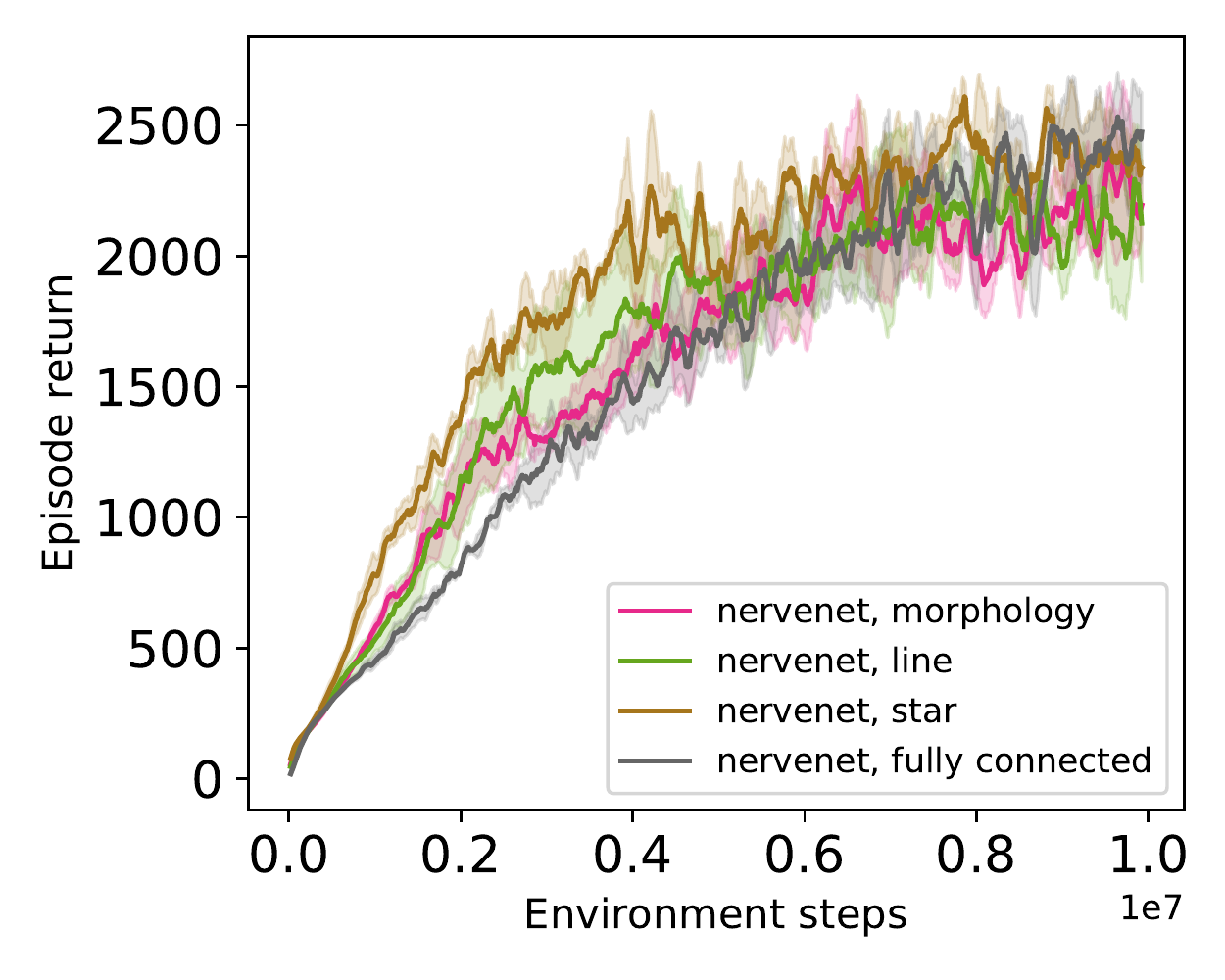}
        \caption{\meth{NerveNet}, \texttt{Walkers}}
        \label{fig:nervenet-structures}
    \end{subfigure}
    \caption{\wbr{Neither \gls{smp} nor \meth{NerveNet}}{\gls{smp}  does not}  leverage the \wbr{agent's }{}morphological information, or the positive effects are outweighted by \wbr{their}{its} negative effect on message passing. 
    }
\end{figure}

To determine if information about the agent's morphology encoded in the
relational graph structure is essential to the success of~\gls{smp}, we compare
its performance given full information about the structure (morphology), given
no information about the structure (star), and given a structural bias unrelated
to the agent's morphology (line).
Ideally, we would test a fully connected architecture as well, but~\gls{smp} only works with trees.
Figure~\ref{fig:topologies} in Appendix~\ref{ref:app-morphology-ablations} illustrates the tested topologies.

The results in Figure~\ref{fig:structure-ablations-a} and~\ref{fig:structure-ablations-b} demonstrate that, surprisingly, performance is not contingent on having information about the physical morphology. A \texttt{star} agent performs on par with the \texttt{morphology} agent, thus refuting the assumption that the method learns because it exploits information about the agent's physical morphology.
The \texttt{line} agent performs worse, perhaps because the network must propagate messages even further away, and information is lost with each hop due to the finite size of the MLPs causing information bottlenecks~\citep{alon2020bottleneck}.

We also present similar results for \meth{NerveNet}.
Figure~\ref{fig:nervenet-structures} shows that all of the variants we tried perform similarly well on \texttt{Walkers} from \citep{wang2018nervenet}, with \texttt{star} being marginally better.
Since \meth{NerveNet} can process non-tree graphs, we also tested a fully connected variant.
This version learns more slowly at the beginning, probably because of difficulties with differentiating nodes at the aggregation step.
Interestingly, in contrast to~\gls{smp}, in \meth{NerveNet}  \texttt{line}  performs on par with \texttt{morphology}.
This might be symptomatic of problems with the message-passing mechanism of~\gls{smp}, e.g., bottlenecks leading to information loss.
\section{\Method{}}
\label{sec:experiments}

Inspired by the results above, we propose \method{}, a transformer-based method for incompatible~\gls{mtrl} in continuous control.
\method{} is motivated by the hypothesis that any benefit \glspl{gnn} can extract from the morphological domain knowledge encoded in the graph is outweighed by the difficulty that the graph creates for message passing.  
In a sparse graph, crucial state information must be communicated across multiple hops, which we hypothesise is difficult to learn in practice.  

\method{} belongs to the encode-process-decode family of architectures~\citep{battaglia2018relational} with a transformer at its core.
Since transformers can be seen as~\glspl{gnn} operating on fully connected graphs, this approach allows us to learn a message passing schema for each state \wbr{and each pass }{}separately, and limits the number of message passes needed to propagate sufficient information through the graph.
Multi-hop message propagation in the presence of aggregation, which could cause
problems with gradient propagation and information loss, is no longer required.
\begin{figure}[b!] %
\centering
\includegraphics[width=.7\textwidth]{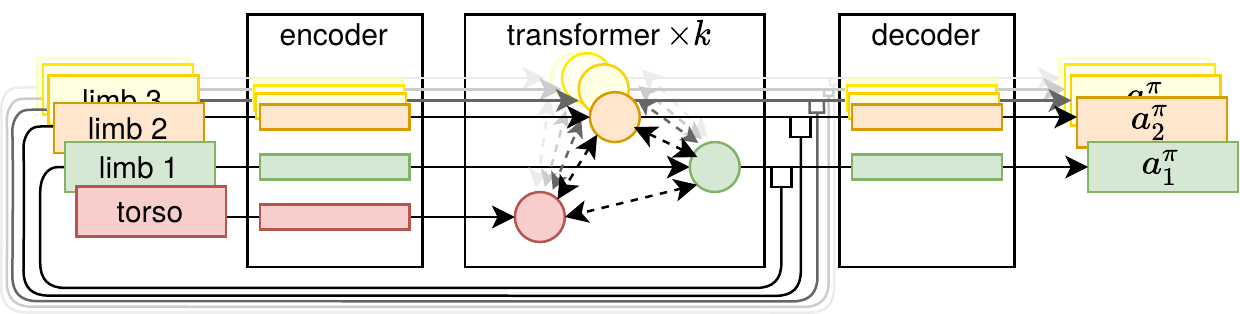}
\caption{\method{} architecture. Lines with squares at the end denote concatenation. Arrows going separately through encoder and decoder denote that rows of the input matrix are processed independently as batch elements. Dashed arrows denote message-passing in a transformer block. The diagram depicts the policy network, the critic has an identical architecture, with the decoder outputs interpreted as value function values.}
\label{fig:amorpheus-diagram}
\end{figure}
We implement both actor and critic in the~\gls{smp}
codebase~\citep{huang2020smp} and made our implementation available online at \url{https://github.com/yobibyte/amorpheus}.
Like in~\gls{smp}, there is no weight sharing between the actor and the critic.
Both of them consist of three parts: a linear encoder, a transformer in the middle, and the output decoder \gls{mlp}.

Figure~\ref{fig:amorpheus-diagram} illustrates the \method{} architecture (policy).
The encoder and decoder process each node independently, as if they are different elements of a mini-batch.
Like~\gls{smp}, the policy network has one output per graph node.
The critic has the same architecture as on Figure~\ref{fig:amorpheus-diagram}, and, as in~\citet{huang2020smp}, each critic node outputs a scalar with the value loss independently computed per node.

Similarly to~\meth{NerveNet} and~\gls{smp}, \method{} is modular and can be used in incompatible environments, including those not seen in training.
In contrast to~\gls{smp} which is constrained by the maximum number of children per node seen at the model initialisation in training, \method{} can be applied to any other morphology with no constraints on the physical connectivity.

Instead of one-hot encoding used in natural language processing, we apply a
linear layer on node observations.
Each node observation uses the same state representation as~\gls{smp} and includes a limb type (e.g. hip or shoulder), position with a relative $x$ coordinate of the limb with respect to the torso, positional and rotational velocities, rotations, angle and possible range of the values for the angle normalised to $[0,1]$.
We add residual connections from the input features to the decoder output to
avoid the nodes forgetting their own features by the time the decoder
independently computes the actions.
Both actor and critic use two attention heads for each of the three transformer layers.
Layer Normalisation~\citep{ba2016layer} is a crucial component of transformers which we also use in \method{}.
See Appendix~\ref{sec:reproducibility} for more details on the implementation.

\subsection{Experimental Results}
\begin{figure}[h]
    \centering
    \begin{subfigure}[t]{0.32\textwidth}
        \centering
        \includegraphics[width=\textwidth]{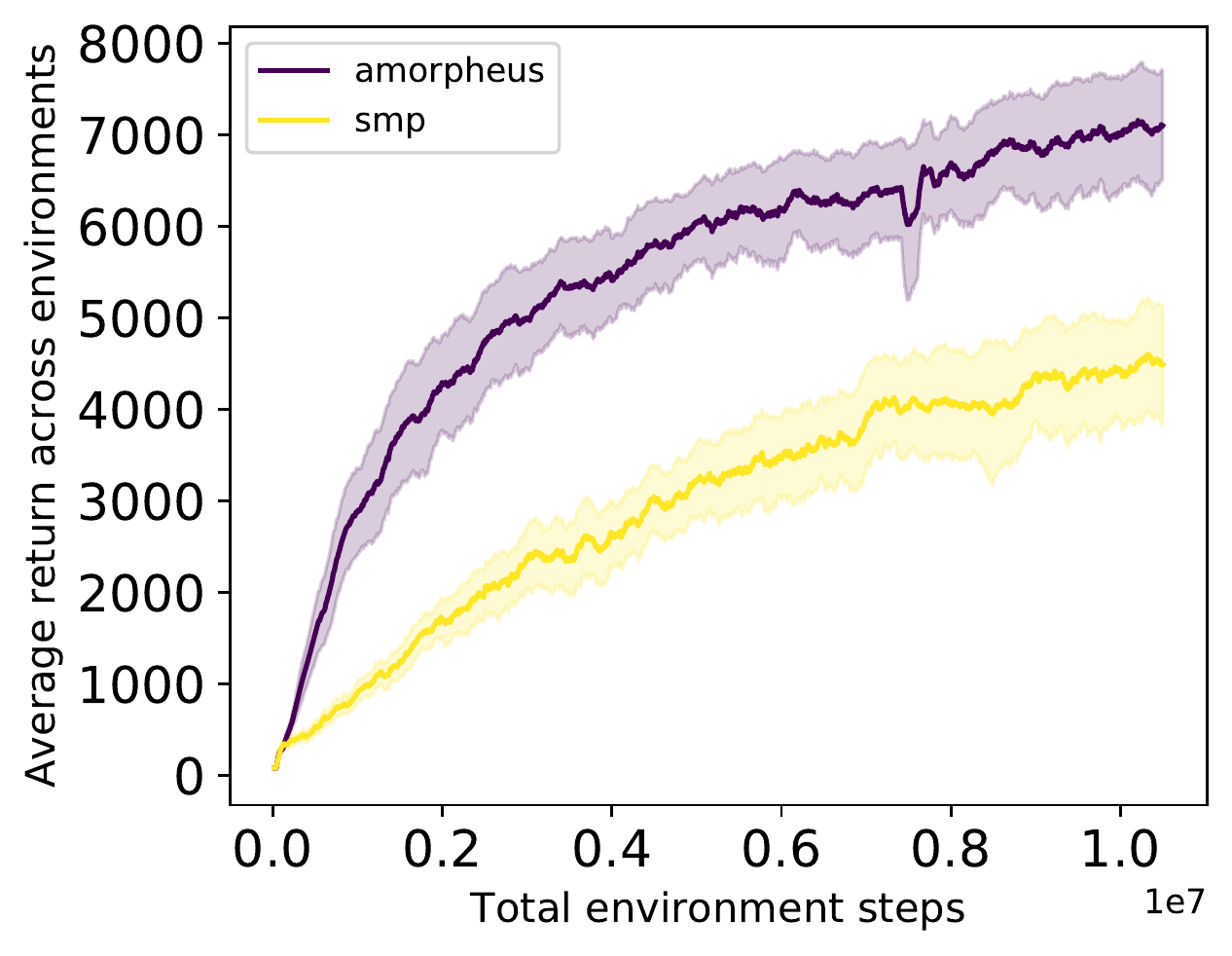}
        \caption{Walker++}
    \end{subfigure}
    \hfill
    \begin{subfigure}[t]{0.32\textwidth}
        \centering
        \includegraphics[width=\textwidth]{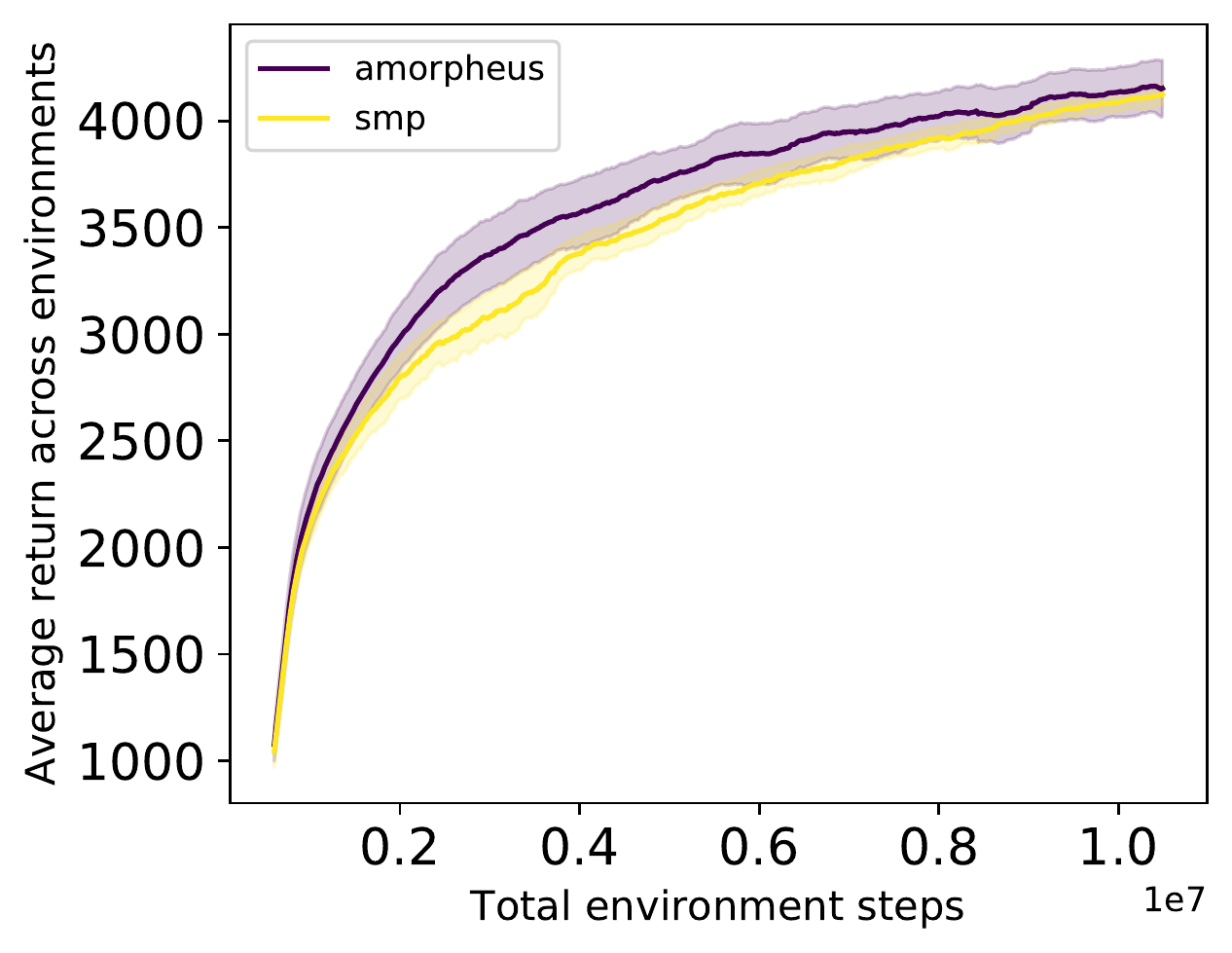}
        \caption{Cheetah++}
    \end{subfigure}
    \hfill
    \begin{subfigure}[t]{0.32\textwidth}
        \centering
        \includegraphics[width=\textwidth]{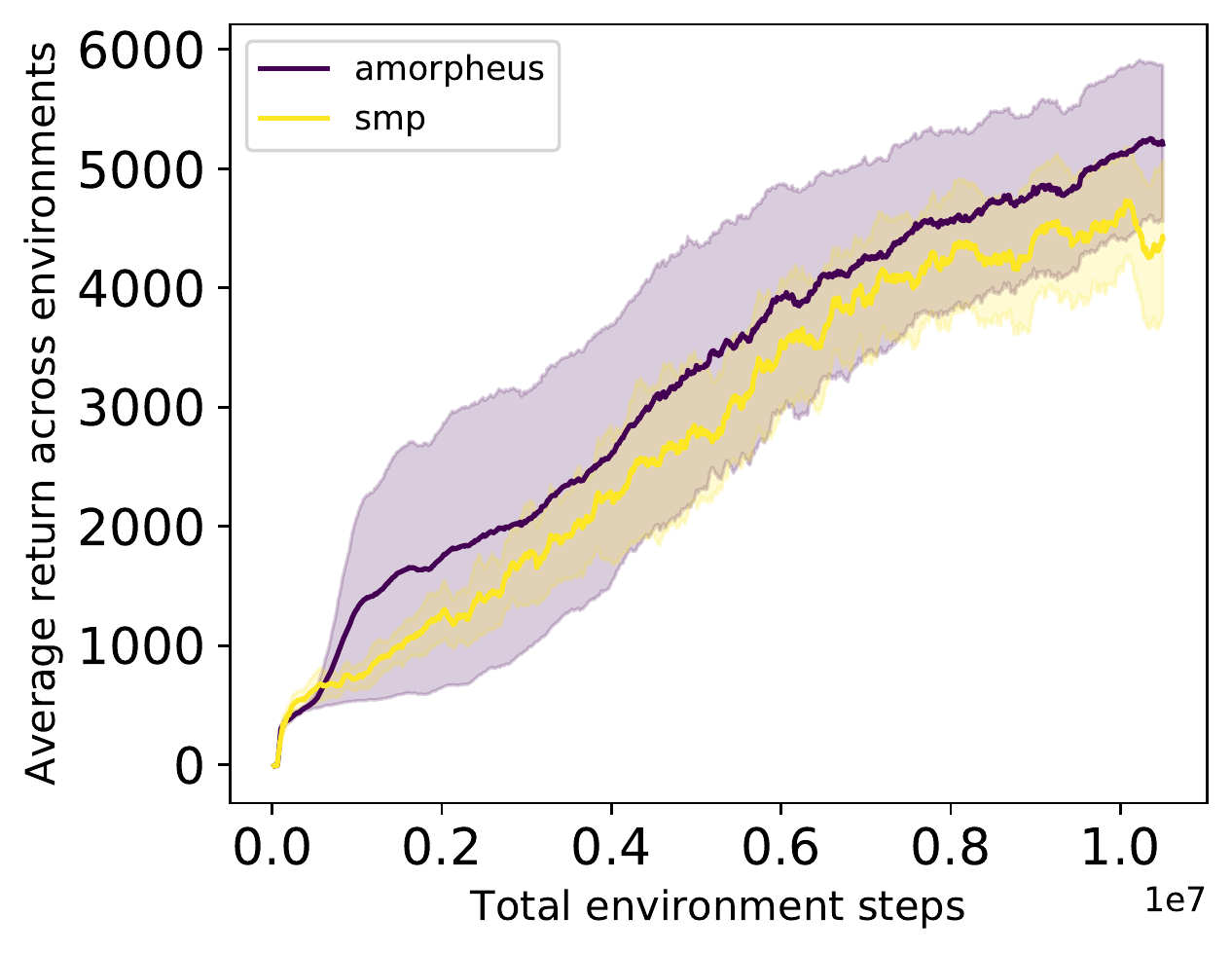}
        \caption{Humanoid++}
    \end{subfigure}
    \hfill
    \begin{subfigure}[t]{0.32\textwidth}
        \centering
        \includegraphics[width=\textwidth]{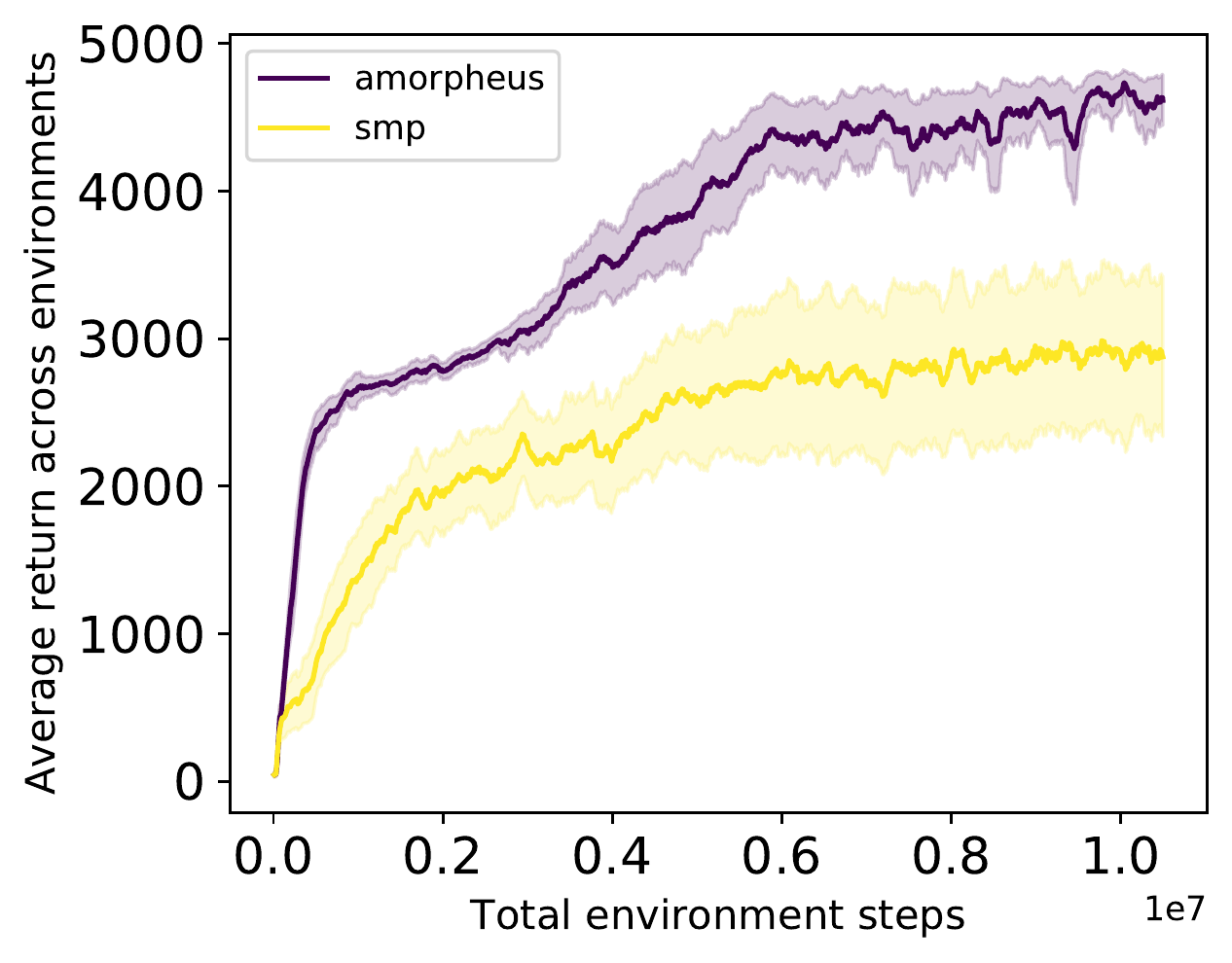}
        \caption{Hopper++}
    \end{subfigure}
    \begin{subfigure}[t]{0.32\textwidth}
        \centering
        \includegraphics[width=\textwidth]{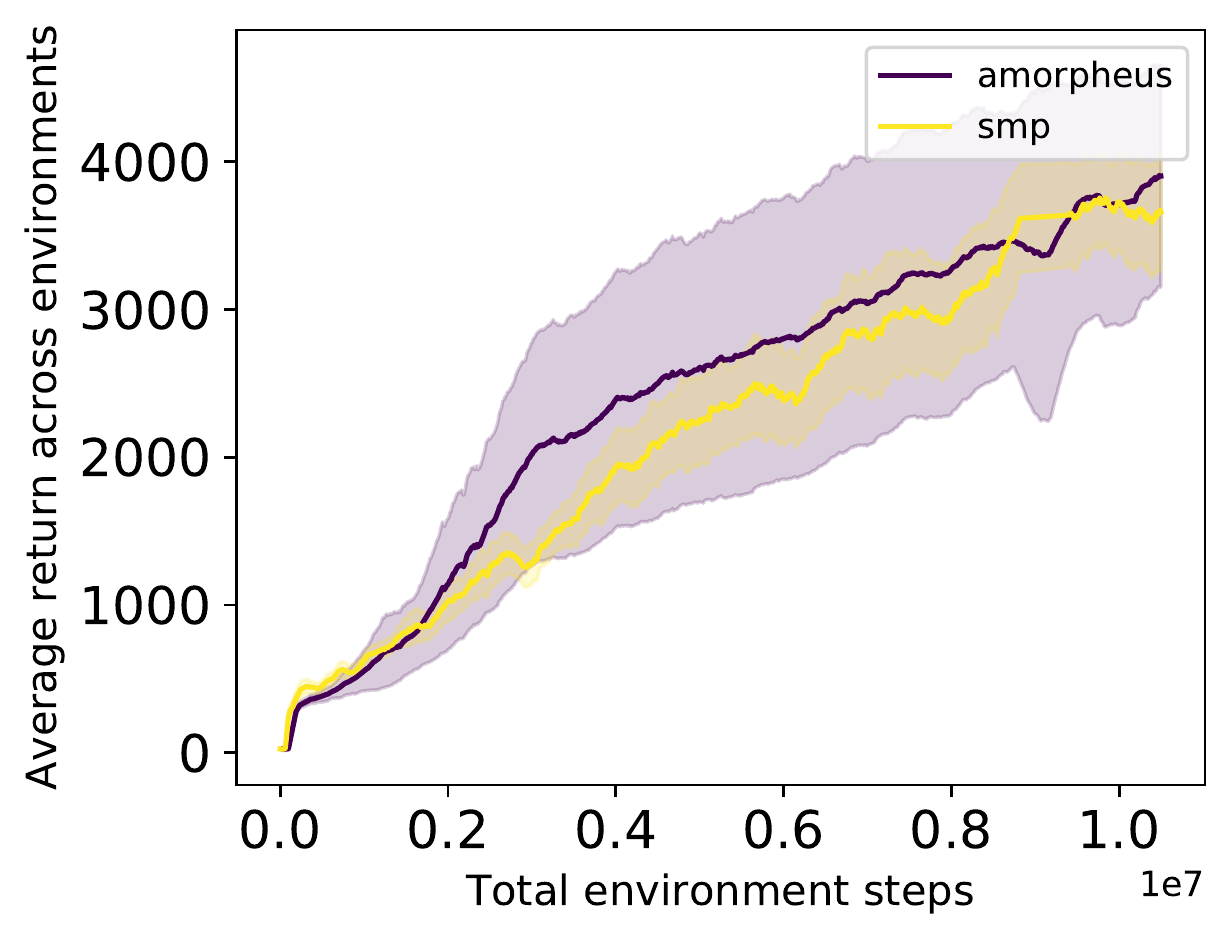}
        \caption{\scriptsize Walker-Humanoid++}
    \end{subfigure}
      \begin{subfigure}[t]{0.32\textwidth}
        \centering
        \includegraphics[width=\textwidth]{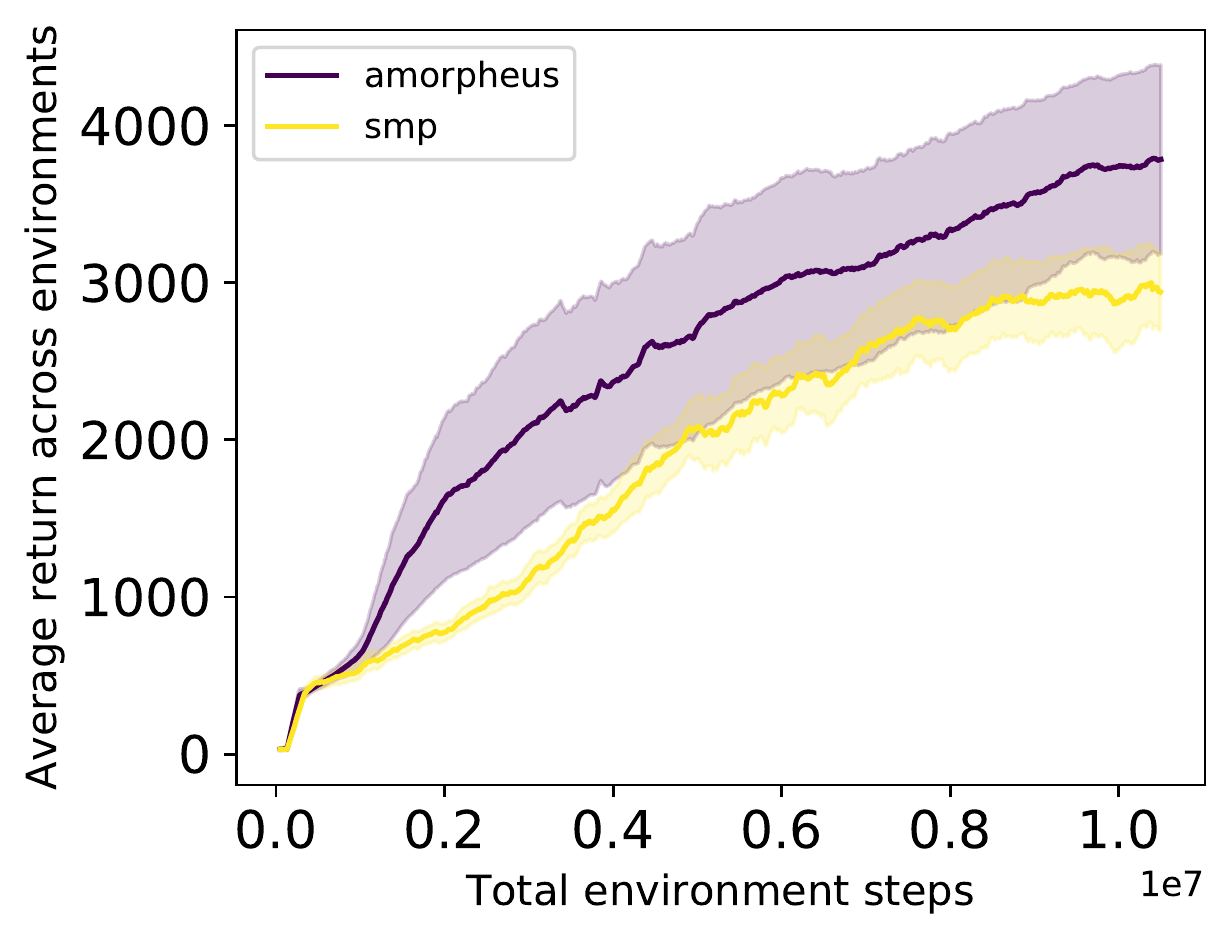}
        \caption{\scriptsize Walker-Humanoid-Hopper++}
    \end{subfigure}
    \begin{subfigure}[t]{0.32\textwidth}
        \centering
        \includegraphics[width=\textwidth]{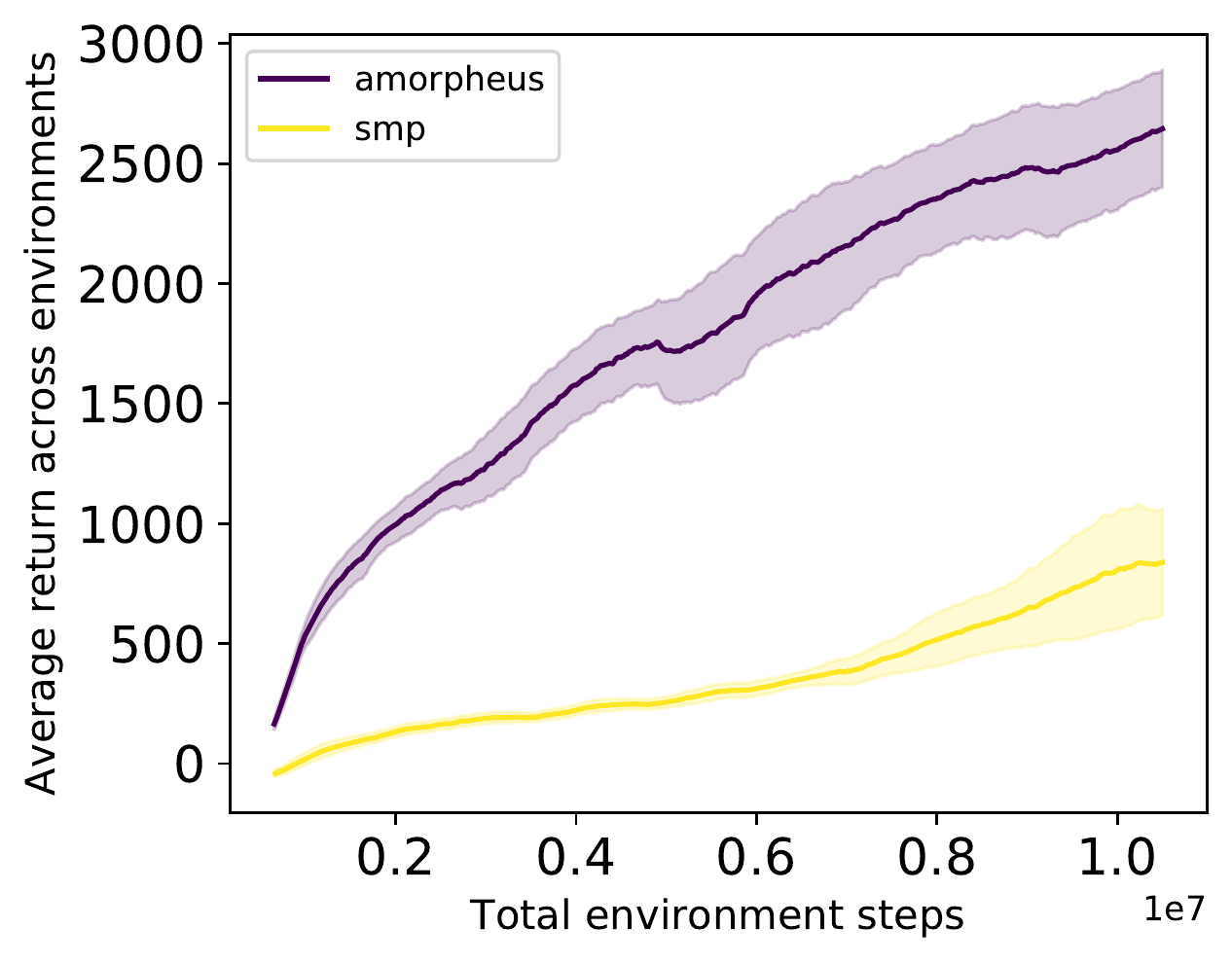}
        \caption{Cheetah-Walker-Humanoid++}
        \label{fig:mtrl-cwh}
    \end{subfigure}
    \begin{subfigure}[t]{0.32\textwidth}
        \centering
        \includegraphics[width=\textwidth]{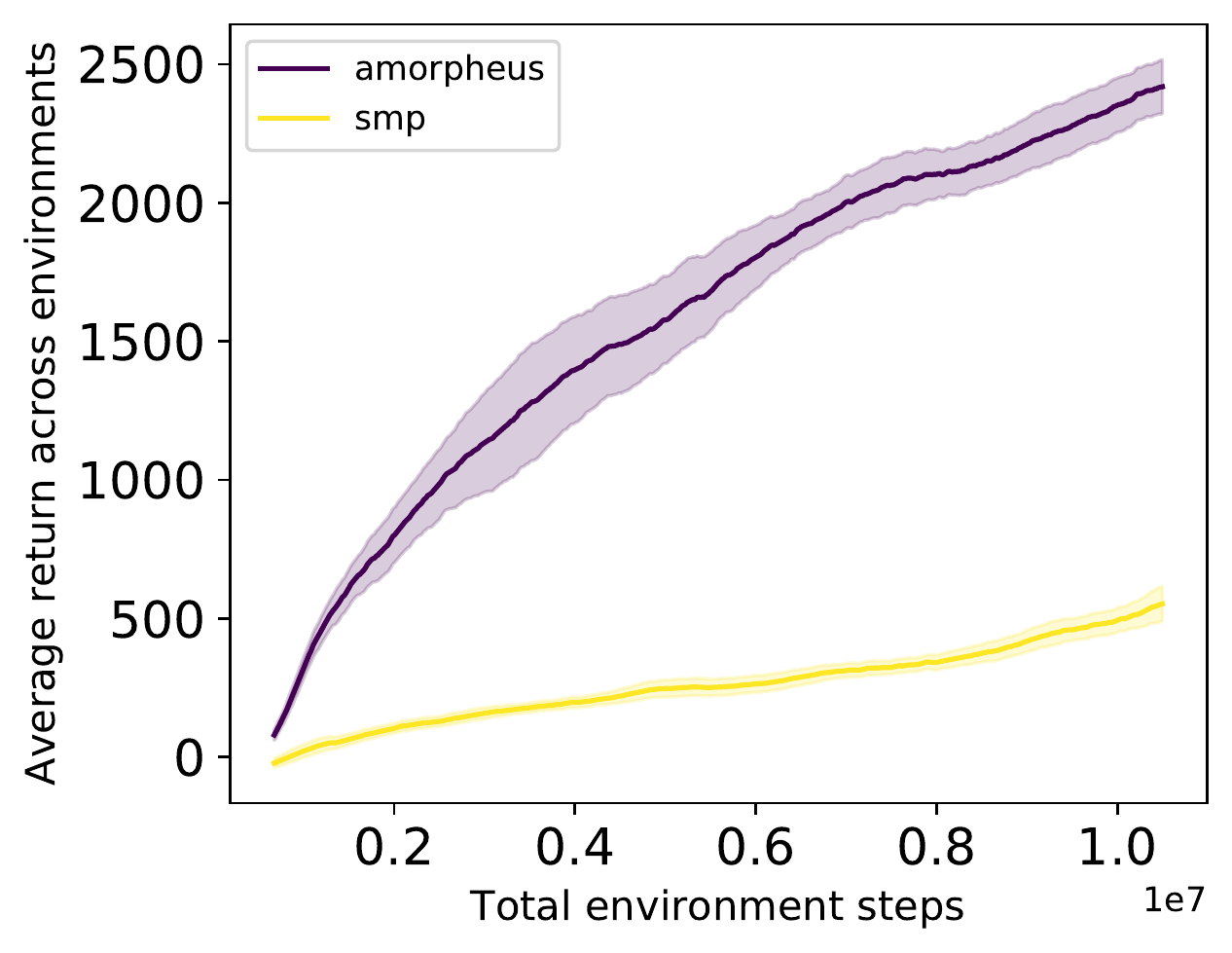}
        \caption{\scriptsize Cheetah-Walker-Humanoid-Hopper++}
        \label{fig:mtrl-cwhh}
    \end{subfigure}
    \caption{\method{} consistently outperforms~\gls{smp} 
    			on \gls{mtrl} benchmarks from \citet{huang2020smp},
    			supporting our hypothesis that no explicit structural information 
    			is needed to learn a successful MTRL policy 
    			and that facilitated message-passing procedure results in faster
          learning.}
    \label{fig:mtrl-on-smp-benchmark}
\end{figure}
We first test \method{} on the set of \gls{mtrl} environments proposed by~\citet{huang2020smp}.
For \texttt{Walker++}, we omit flipped environments, since~\citet{huang2020smp} implement flipping on the model level.
For \method{}, the flipped environments look identical to the original ones.
Our experiments in this Section are built on top of the \meth{TD3} implementation used in~\citet{huang2020smp}.

Figure~\ref{fig:mtrl-on-smp-benchmark} supports our hypothesis that explicit morphological information encoded in graph topology is not needed to yield a single policy achieving high average returns across a set of incompatible continuous control environments.
Free from the need to learn multi-hop communication and equipped with the attention mechanism, \method{} clearly outperforms~\gls{smp}, the state-of-the-art algorithm for incompatible continuous control.
\citet{huang2020smp} report that training \gls{smp} on \texttt{Cheetah++} together with other environments makes~\gls{smp} unstable. By contrast, \method{} has no trouble learning in this regime (Figure~\ref{fig:mtrl-cwh} and~\ref{fig:mtrl-cwhh}).

Our experiments demonstrate that node features have enough information for \method{} to perform the task and limb discrimination needed for successful~\gls{mtrl} continuous control policies.
For example, a model can distinguish left from right, not from structural biases as in~\gls{smp}, but from the relative position of the limb w.r.t.\ the root node provided in the node features.

While the total number of tasks in the~\gls{smp} benchmarks is high, they all share one key characteristic.
All tasks in a benchmark are built using subsets of the limbs from an archetype (e.g., \texttt{Walker++} or \texttt{Cheetah++}).
To verify that our results hold more broadly, we adapted the \texttt{Walkers}
benchmark \citep{wang2018nervenet} and compared \method{} with \gls{smp} and
\meth{NerveNet} on it.
This benchmark includes five agents with different morphologies: a Hopper, a HalfCheetah, a FullCheetah, a Walker, and an Ostrich.
The results in Figure~\ref{fig:nervenet-walkers-mtrl} are consistent\footnote{
\wbr{Note that}{However,} the performance of \meth{NerveNet} is not directly comparable, 
as the observational features and the learning algorithm differ from 
\method{} and \gls{smp}.  We do not test \meth{NerveNet} on \gls{smp} benchmarks
because the codebases are not compatible and comparing \meth{NerveNet}
and~\gls{smp} is not the focus of the paper.
Even if we implemented \meth{NerveNet} in the \gls{smp} training loop, it is unclear how the critic of \meth{NerveNet} would perform in a new setting.
The original paper considers two options for the critic: one GNN-based and one MLP-based. We use the latter in Figure~\ref{fig:nervenet-walkers-mtrl} as the former takes only the root node output labels as an input and is thus most likely to face difficulty in learning multi-hop message-passing.
The MLP critic should perform better because training an MLP is easier, though it might be sample-inefficient when the number of tasks is large.
For example, in \texttt{Cheetah++} an agent would need to learn 12 different critics.
Finally, \meth{NerveNet} learns a separate MLP encoder per task, partially defeating the purpose of using \gls{gnn} for incompatible environments.
} with our previous experiments, demonstrating the benefits of \method' fully-connected graph with attentional aggregation.

While we focused on~\gls{mtrl} in this work, we also evaluated \method{} in a zero-shot generalisation setting.
Table~\ref{tab:generalisation-results} in Appendix~\ref{sec:generalisation-results} provides initial results demonstrating \method{}'s potential.

\subsection{Attention Mask Analysis}

\gls{gnn}-based policies, especially those that use attention, are more interpretable than monolithic \gls{mlp} policies.
We now analyse the attention masks that \method{} learns.
\begin{figure}[t]
    \begin{minipage}[t]{.33\textwidth}
        \centering
        \includegraphics[height=88pt]{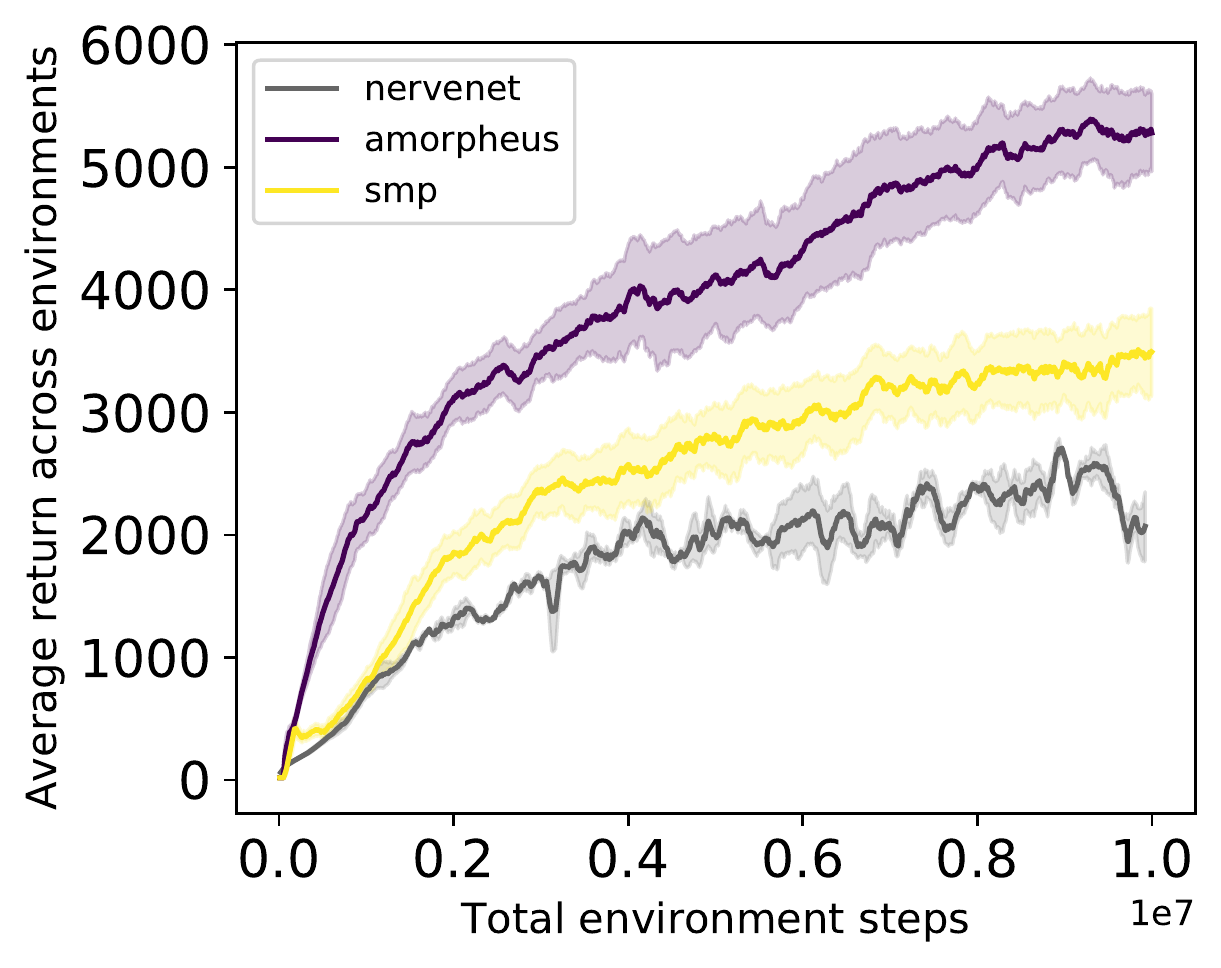}
        \caption{\gls{mtrl} performance on \texttt{Walkers}~\citep{wang2018nervenet}.}
        \label{fig:nervenet-walkers-mtrl}
    \end{minipage}
    \hfill
    \begin{minipage}[t]{0.64\textwidth}
        \centering
        \includegraphics[width=\textwidth]{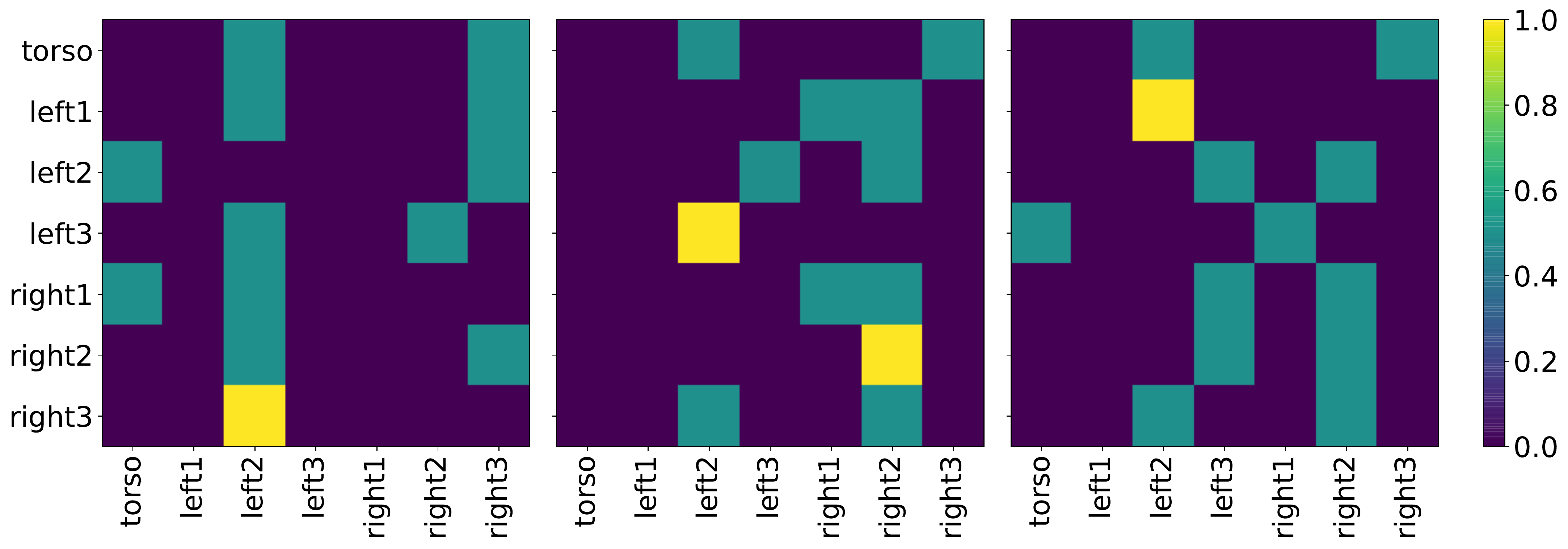}
        \caption{State-dependent masks of \method{} (3\textsuperscript{rd} attention layer) within a \texttt{Walker}-\texttt{7} rollout.}
        \label{fig:mask-variety}
    \end{minipage}
    \begin{minipage}[t]{\textwidth}
        \centering
        \includegraphics[width=0.9\textwidth]{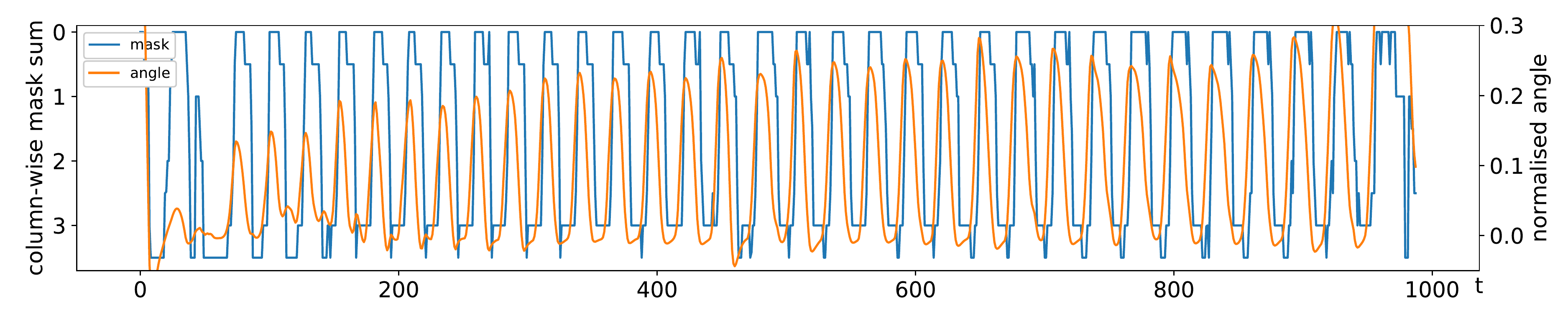}
        \caption{In \wbr{the first attention layer of a }{}\texttt{Walker}-\texttt{7}\wbr{~rollout}{, in the first layer}, nodes attend to an upper leg \wbr{(column-wise mask sum $\sim$ 3) }{}when the leg is closer to the ground (\wbr{normalized }{}angle $\sim$ 0).} 
        \label{fig:mask_cycles}
    \end{minipage}
\end{figure}
Having an implicit structure that is state dependent is one of the benefits of \method{} (every node has access to other nodes' annotations, and the aggregation weights depend on the input as shown in Equation~\ref{eq:attention}).
By contrast, \meth{NerveNet} and~\gls{smp} have a rigid message-passing structure that does not change throughout training or throughout a rollout.
Indeed, Figure~\ref{fig:mask-variety} shows a variety of masks a \texttt{Walker++} model exhibits within a \texttt{Walker}-\texttt{7} rollout, confirming that \method{} attends to different parts of the state space based on the input.

Both~\citet{wang2018nervenet} and~\citet{huang2020smp} notice periodic patterns arising in their models.
Smilarly, \method{} demonstrates cycles in attention masks, usually arising for the first layer of the transformer.
Figure~\ref{fig:mask_cycles} shows the column-wise sum of the attention masks coordinated with an upper-leg limb of a \texttt{Walker}-\texttt{7} agent.
Intuitively, the column-wise sum shows how \wbr{much }{}other nodes are interested in the node corresponding to that column.

Interestingly, attention masks in earlier layers change more slowly within a rollout than those of the downstream layers.
Figure~\ref{fig:walker-cumulative-change} in Appendix~\ref{sec:mask-cumulative-change} demonstrates this phenomenon for three different \texttt{Walker++} models tested on \texttt{Walker-7}.
This shows that \method{} might, in principle, learn a rigid structure (as in~\glspl{gnn}) if needed.

Finally, we investigate how attention masks evolve over time.
Early in training, the masks are spread across the whole graph.
Later on, the mask weights distributions become less uniform.
Figures~\ref{fig:mask-evolution-1},~\ref{fig:mask-evolution-2} and \ref{fig:mask-evolution-3} in Appendix~\ref{sec:mask-evolution} demonstrate this phenomenon on \texttt{Walker}-\texttt{7}.
\section{Related Work}

Most~\gls{mtrl} research considers the compatible case~\citep{rusu2015policy, parisotto2015actor, NIPS2017_7036, varghese2020mtrl}.
\gls{mtrl} for continuous control is often done from pixels with CNNs solving part of the compatibility issue.
DMLab~\citep{beattie2016deepmind} is a popular choice when learning from pixels with  a compatible action space shared across the environments~\citep{hessel2019multi, song2019v}.

\glspl{gnn} started to stretch the possibilities of RL allowing~\gls{mtrl} in incompatible environments.
\citet{khalil2017learning} learn combinatorial optimisation algorithms over graphs.
\citet{kurin2019improving} learn a branching heuristic of a SAT solver.
Applying approximations schemes typically used in~\gls{rl} to these settings is impossible, because they expect input and output to be of fixed size.
Another form of (potentially incompatible)~\gls{rl} using message passing are 
coordination graphs \citep[e.g.~DCG,][]{boehmer2020dcg}, 
that use the max-plus algorithm \citep{pearl88prob}
to coordinate action selection between multiple agents.
One can apply DCG in single-agent~\gls{rl} using ideas of \citet{tavakoli2021learning}.

Several methods for incompatible continuous control have also been proposed.
\citet{chen2018hardware} pad the state vector with zeros to have the same dimensionality for robots with different number of joints, and condition the policy on the hardware information of the agent.
\citet{d2019sharing} demonstrate a positive effect of learning a common network for multiple tasks, learning a specific encoder and a decoder one per task. 
We expect this method to suffer from sample-inefficiency because it has to learn separate input and output heads per each task.
Moreover, \citet{wang2018nervenet} have a similar implementation of their~\gls{mtrl} baseline showing that~\glspl{gnn} have benefits over MLPs for incompatible control.
\citet{huang2020smp}, whose work is the main baseline in this paper, apply a~\gls{gnn}-like approach and study its~\gls{mtrl} and generalisation properties.
The method can be used only with trees, its aggregation function is not permutation invariant, and the message-passing schema stays fixed throughout the training procedure.
\citet{wang2018nervenet} and~\citet{huang2020smp} attribute the effectiveness of their methods to the ability of the~\glspl{gnn} to exploit information about agent morphology.
In this work, we present evidence against this hypothesis, showing that existing approaches do not exploit morphological information as was previously believed.

Attention mechanisms have also been used in the~\gls{rl} setting.
\citet{zambaldi2018relational} consider self-attention to deal with an object-oriented state space. 
They further generalize this to variable action spaces and test generalisation on Starcraft-II mini-games that have a varying number of units and other environmental entities.
\citet{duan2017one} apply attention for both temporal dependency and a factorised state space (different objects in the scene) keeping the action space compatible.
\citet{parisotto2019stabilizing} use transformers as a replacement for a recurrent policy.
\citet{icml2020_1696} use transformers to add history dependence in a POMDP as well as for factored observations, having a node per game object.
The authors do not consider a factored action space, with the policy receiving the aggregated information of the graph after the message passing ends.
\citet{baker2019emergent} use self-attention to account for a factored state-space to attend over objects or other agents in the scene.
\method{} does not use a transformer for recurrency but for the factored state and action spaces, with each non-torso node having an action output.
\citet{iqbal2019maac} apply attention to generalise \gls{mtrl} multi-agent policies over varying environmental objects and \citet{iqbal2020aiqmix} extend this to a factored action space by summarising the values of all agents with a mixing network \citep{rashid2018qmix}.
\citet{li2020deep} learn embeddings for a multi-agent actor-critic architecture by generating the weights of a graph convolutional network \citep[GCN,][]{kipf2016semi} with attention. This allows a different topology in every state, similar to \method, which goes one step further and allows to change the topology in every round of message passing.

Another line of work aims to infer graph topology instead of hardcoding one.
Differentiable Graph Module~\citep{DBLP:journals/corr/abs-2002-04999} predicts edge probabilities doing a continuous relaxation of k-nearest neighbours to differentiate the output with respect to the edges in the graph.
\citet{DBLP:conf/nips/0003LT20} learn to augment a given graph with additional edges to improve the performance of a downstream task.
\citet{DBLP:conf/icml/KipfFWWZ18} use variational autoencoders~\citep{DBLP:journals/corr/KingmaW13} using a~\gls{gnn} for reconstruction.
Notably, the authors notice that message passing on a fully connected graph might work better than when restricted by skeleton when evaluated on human motion capture data.
\section{Conclusions and Future Work}
In this paper, we investigated the role of explicit morphological information in graph-based continous control.
We ablated existing methods~\gls{smp} and \meth{NerveNet}, providing evidence against the belief that these methods improve performance by exploiting explicit morphological structure encoded in graph edges. 
Motivated by our findings, we presented \method{}, a transformer-based method for~\gls{mtrl} in incompatible environments.
\method{} obviates the need to propagate messages far away in the graph and can attend to different regions of the observations depending on the input and the particular point in training.
As a result, \method{} clearly outperforms existing work in incompatible continuous control.
In addition, \method{} exhibits non-trivial behaviour such as periodic cycles of attention masks coordinated with the gait.
The results show that information in the node features alone is enough to learn a successful~\gls{mtrl} policy.
We believe our results further push the boundaries of incompatible~\gls{mtrl} and provide valuable insights for further progress.

One possible drawback of~\method{} is its computational complexity.
Transformers suffer from quadratic complexity in the number of nodes with a growing body of work addressing this issue~\citep{tay2020efficient}.
However, the number of the nodes in continuous control problems is relatively low compared to much longer sequences used in NLP~\citep{devlin2018bert}.
Moreover, Transformers are higly parallelisable, compared to~\gls{smp} with the data dependency across tree levels (the tree is processed level by level with each level taking the output of the previous level as an input).

We focused on investigating the effect of injecting explicit morphological information into the model.
However, there are also opportunities to improve the learning algorithm itself.
Potential directions of improvement include averaging gradients instead of performing sequential task updates, or balancing tasks updates with multi-armed bandits or PopArt~\citep{hessel2019multi}.
\subsubsection*{Acknowledgments}

VK is a doctoral student at the University of Oxford funded by Samsung R\&D Institute UK through the AIMS program.
SW has received funding from the European Research Council under the European Union’s Horizon 2020 research and innovation programme (grant agreement number 637713).
The experiments were made possible by a generous equipment grant from NVIDIA.
The authors would like to thank Henry Kenlay and Marc Brockschmidt for useful discussions on~\glspl{gnn}.
\bibliography{iclr2021_conference}
\bibliographystyle{iclr2021_conference}

\clearpage
\appendix
\section{Reproducibility}
\label{sec:reproducibility}

We initially took the transformer implementation from the Official Pytorch  Tutorial~\citep{pytorchTransformerTutorial} which uses \texttt{TransformerEncoderLayer} from
Pytorch~\citep{paszke2017automatic}.
We modified it for the regression task instead of classification, and removed masking and the positional encoding.
Table~\ref{tab:hyperparameters} provides all the hyperparameters needed to replicate our experiments.

\begin{table*}[h]
    \centering
    \caption{Hyperparameters of our experiments}
    \label{tab:hyperparameters}
    \begin{tabular}{lll}
    \textbf{Hyperparameter}&\textbf{Value}&\textbf{Comment}\\
    \midrule
    \textit{\method{}}&&\\
    \midrule
    -- Learning rate & 0.0001&\\
    -- Gradient clipping & 0.1&\\
    -- Normalisation & LayerNorm & As an argument to \texttt{TransformerEncoder} in \texttt{torch.nn}\\
    -- Attention layers & 3&\\
    -- Attention heads & 2&\\
    -- Attention hidden size & 256&\\
    -- Encoder output size & 128&\\
    \midrule
    Training&&\\
    \midrule
    -- runs &3& per benchmark\\
    \end{tabular}
\end{table*}

\begin{table*}
    \centering
    \caption{Full list of environments used in this work.}
    \label{tab:environments}
    \begin{tabular}{lll}
    \textbf{Environment}&\textbf{Training}&\textbf{Zero-shot testing}\\
    \midrule
    \texttt{Walker++}&&\\
    \midrule
     & \texttt{walker\_2\_main} &\texttt{walker\_3\_main}\\
     & \texttt{walker\_4\_main} &\texttt{walker\_6\_main}\\
     & \texttt{walker\_5\_main} &\\
     & \texttt{walker\_7\_main} &\\
     \midrule
     \texttt{humanoid++}&&\\
     \midrule
     & \texttt{humanoid\_2d\_7\_left\_arm} &\texttt{humanoid\_2d\_7\_left\_leg}\\
     & \texttt{humanoid\_2d\_7\_lower\_arms} & \texttt{humanoid\_2d\_8\_right\_knee}\\
     & \texttt{humanoid\_2d\_7\_right\_arm} &\\
     & \texttt{humanoid\_2d\_7\_right\_leg} &\\
     & \texttt{humanoid\_2d\_8\_left\_knee} &\\
     & \texttt{humanoid\_2d\_9\_full} &\\
     \midrule
     \texttt{Cheetah++}&&\\
     \midrule
     & \texttt{cheetah\_2\_back} & \texttt{cheetah\_3\_balanced}\\
     & \texttt{cheetah\_2\_front} &\texttt{cheetah\_5\_back}\\
     & \texttt{cheetah\_3\_back} &\texttt{cheetah\_6\_front}\\
     & \texttt{cheetah\_3\_front} &\\
     & \texttt{cheetah\_4\_allback} &\\
     & \texttt{cheetah\_4\_allfront} &\\
     & \texttt{cheetah\_4\_back} &\\
     & \texttt{cheetah\_4\_front} &\\
     & \texttt{cheetah\_5\_balanced} &\\
     & \texttt{cheetah\_5\_front} &\\
     & \texttt{cheetah\_6\_back} &\\
     & \texttt{cheetah\_7\_full} &\\
     \midrule
     \texttt{Cheetah-Walker-}&&\\
     \texttt{-Humanoid}&&\\
     \midrule
     &All in the column above&All in the column above\\
     \midrule
     \texttt{Hopper++}&&\\
     \midrule
     &\texttt{hopper\_3}&\\
     &\texttt{hopper\_4}&\\
     &\texttt{hopper\_5}&\\
     \midrule
     \texttt{Cheetah-Walker-}&&\\
     \texttt{-Humanoid-Hopper}&&\\
     \midrule
     &All in the column above&All in the column above\\
    \midrule
    \texttt{Walkers} from &&\\
    \citet{wang2018nervenet}&&\\
    \midrule
    &\texttt{Ostrich}&\\
    &\texttt{HalfCheetah}&\\
    &\texttt{FullCheetah}&\\
    &\texttt{Hopper}&\\
    &\texttt{HalfHumanoid}&\\
    \end{tabular}
\end{table*}

\begin{figure}[h]
  \centering
  \begin{minipage}[t]{.49\textwidth}
    \centering
    \includegraphics[width=0.8\textwidth]{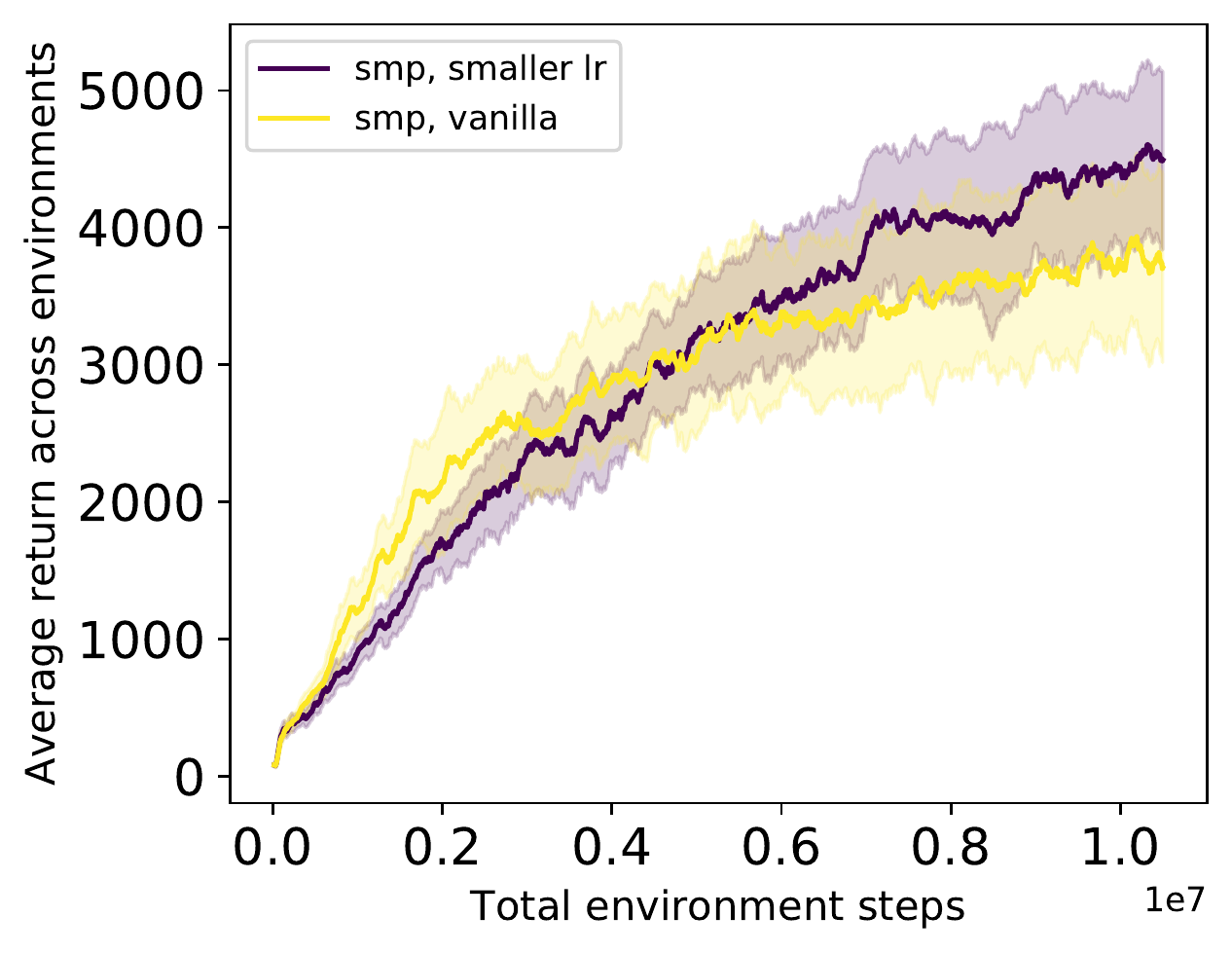}
    \caption{Smaller learning rate make~\gls{smp} to yield better results on \texttt{Walker++}.}
    \label{fig:smp-hyperparameters}
  \end{minipage}
  \hfill
  \begin{minipage}[t]{.49\textwidth}
    \centering
    \includegraphics[width=0.8\textwidth]{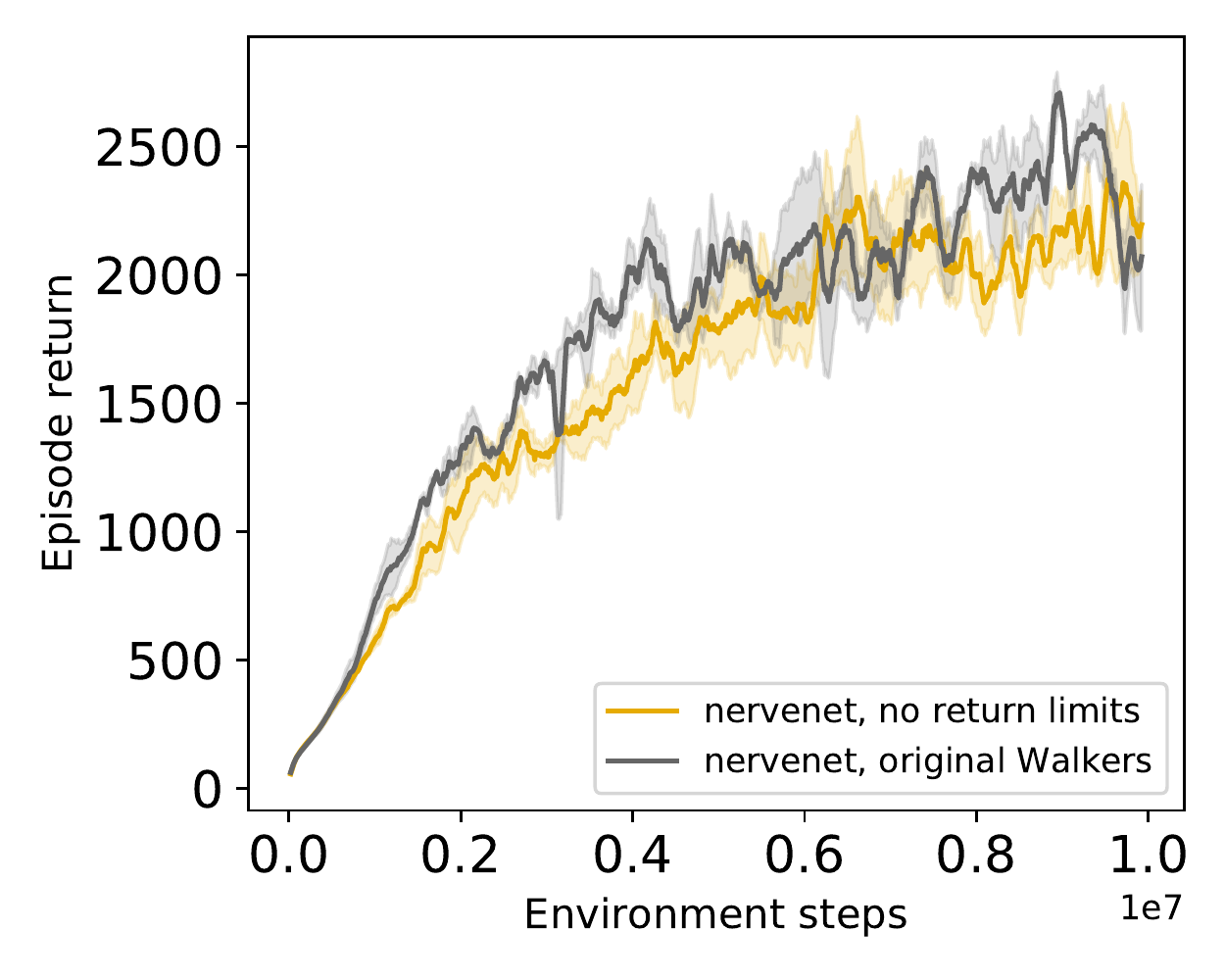}
    \caption{Removing the return limit slightly deteriorates the performance of NerveNet on Walkers.}
    \label{fig:nervenet-walkers-w-wo-limit}
  \end{minipage}
\end{figure}

\method{} makes use of gradient clipping and a smaller learning rate.
We found, that~\gls{smp} also performs better with the decreased learning rate ($0.0001$) as well and we use it throughout the work.
Figure~\ref{fig:smp-hyperparameters} demonstrates the effect of a smaller learning rate on \texttt{Walker++}.
All other \gls{smp} hyperparameters are as reported in the original paper with the two-directional message passing. 

\citet{wang2018nervenet} add an artificial return limit of 3800 for their Walkers environment. 
We remove this limit and compare the methods without it. 
For NerveNet, we plot the results with the option best for it. 
Figure~\ref{fig:nervenet-walkers-w-wo-limit} compares the two options.
\clearpage

\section{Morphology ablations}
\label{ref:app-morphology-ablations}

\begin{figure}[h]
    \centering
    \begin{subfigure}[b]{0.3\textwidth}
        \centering
        \includegraphics[width=\textwidth]{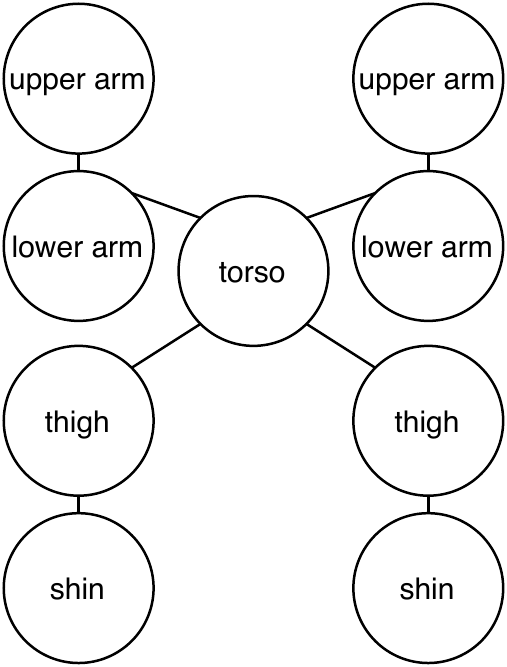}
        \caption{Morphology}
        \label{fig:topologies-morphology}
    \end{subfigure}
    \hfill
    \begin{subfigure}[b]{0.3\textwidth}
        \centering
        \includegraphics[width=\textwidth]{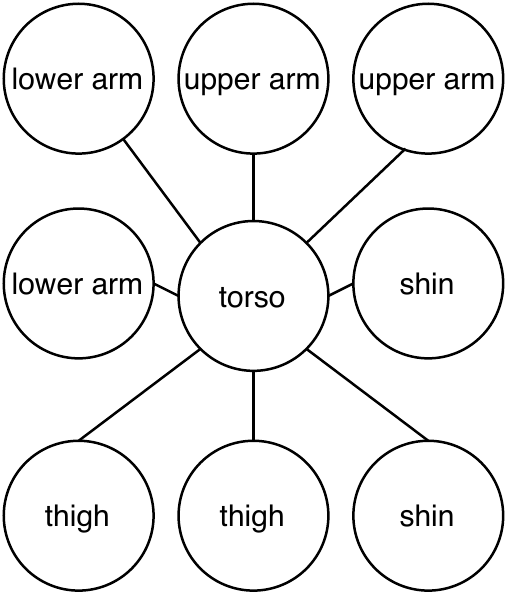}
        \caption{Star}
    \end{subfigure}
    \hfill
    \begin{subfigure}[b]{0.3\textwidth}
        \centering
        \includegraphics[width=\textwidth]{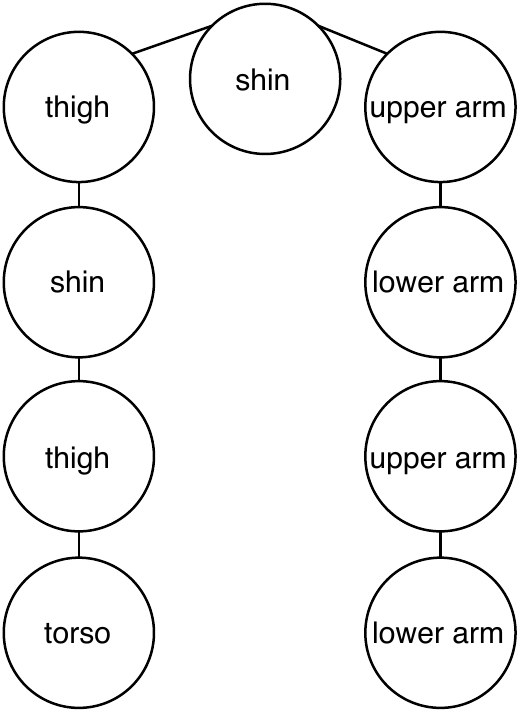}
        \caption{Line}
    \end{subfigure}
    \hfill
    \caption{Examples of graph topologies used in the structure ablation experiments.}
    \label{fig:topologies}
\end{figure}

Figure~\ref{fig:topologies} shows examples of graph topologies we used in structure ablation experiments.

\clearpage
\section{Attention Mask Analysis}

\subsection{Evolution of masks throughout the training process}
\begin{figure}[h]
  \centering
  \includegraphics[width=\textwidth]{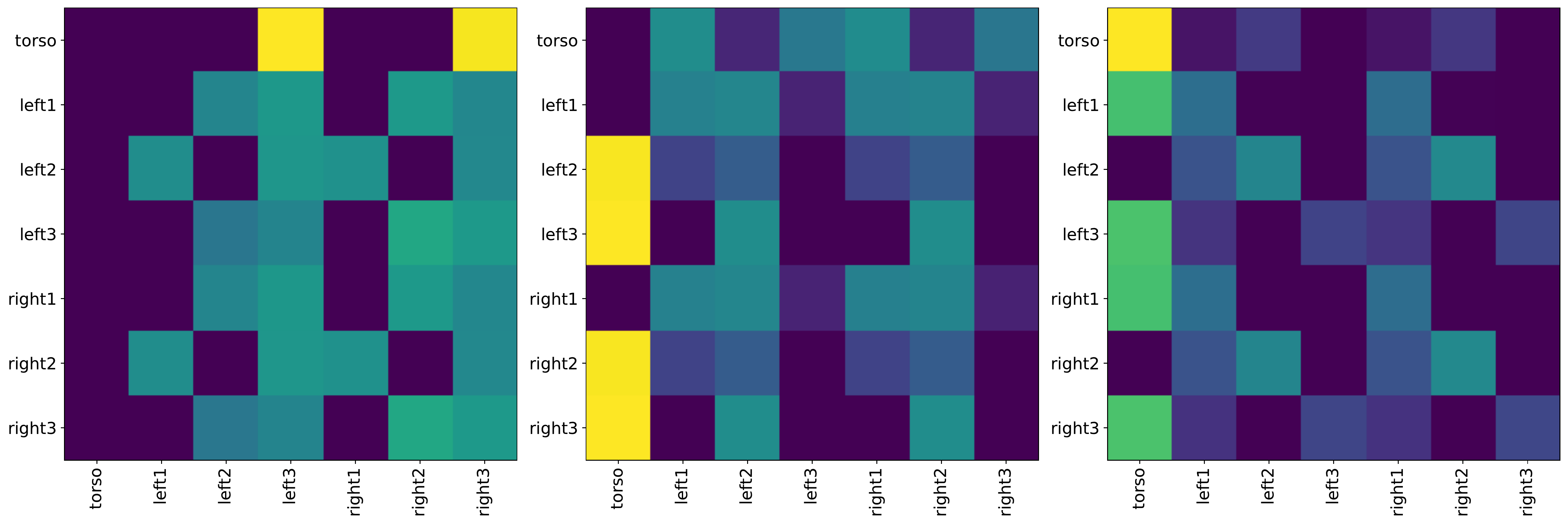}
  \caption{\texttt{Walker++} masks for the 3 attention layers on \texttt{Walker}-\texttt{7} at the beginning of training.}
  \label{fig:mask-evolution-1}
\end{figure}

\begin{figure}[h]
  \centering
  \includegraphics[width=\textwidth]{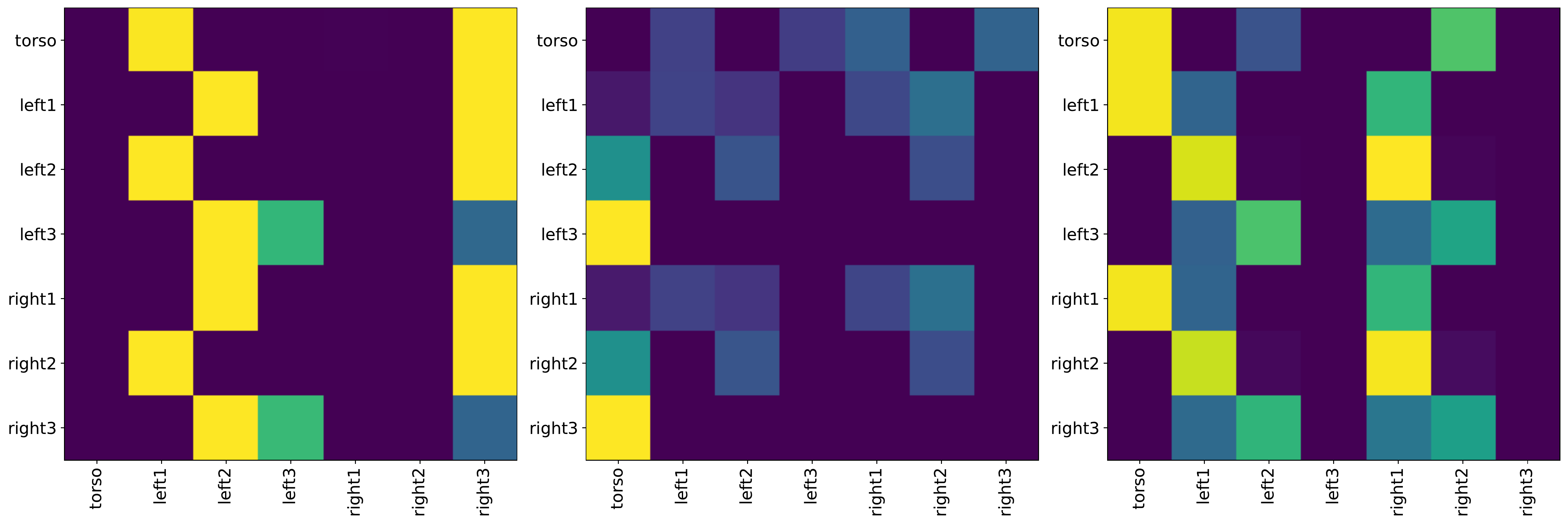}
  \caption{\texttt{Walker++} masks for the 3 attention layers on \texttt{Walker}-\texttt{7} after 2.5 mil frames.}
  \label{fig:mask-evolution-2}
\end{figure}

\begin{figure}[h]
  \centering
  \includegraphics[width=\textwidth]{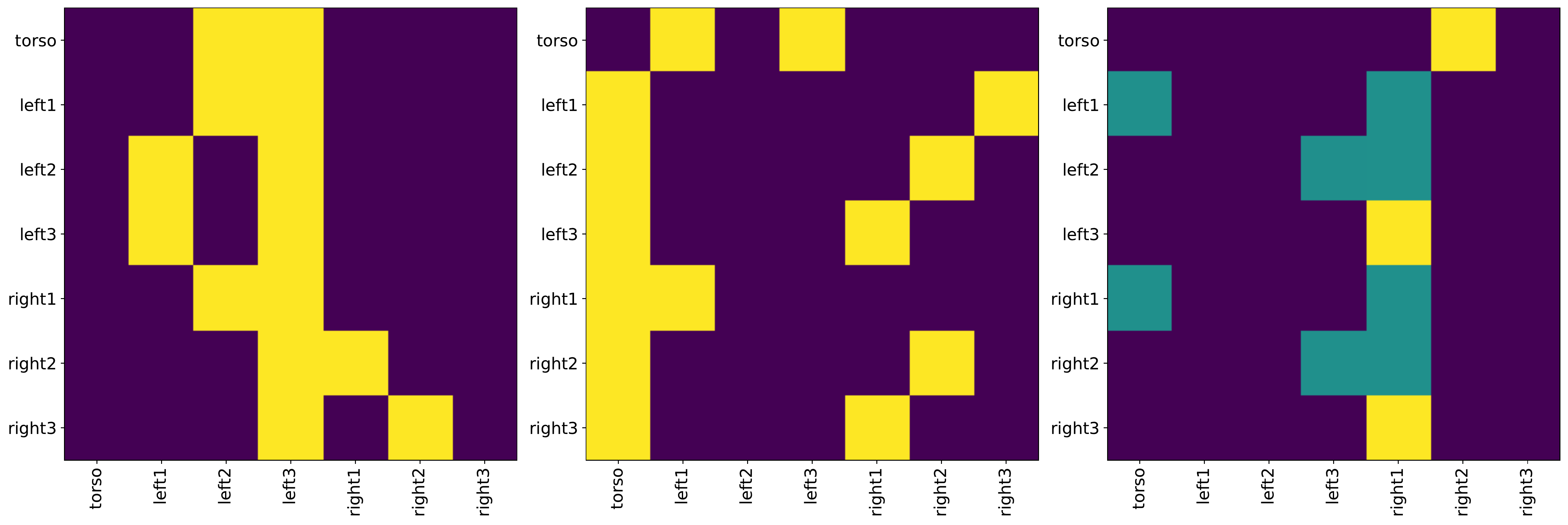}
  \caption{\texttt{Walker++} masks for the 3 attention layers on \texttt{Walker}-\texttt{7} at the end of training.}
  \label{fig:mask-evolution-3}
\end{figure}

Figures \ref{fig:mask-evolution-1}, \ref{fig:mask-evolution-2}
and \ref{fig:mask-evolution-3} demonstrate the evolution 
of \method~attention masks during training.
\label{sec:mask-evolution}
\clearpage
\subsection{Attention masks cumulative change}
\label{sec:mask-cumulative-change}

\begin{figure}[h]
  \centering
  \begin{subfigure}[t]{0.325\textwidth}
      \centering
      \includegraphics[width=\textwidth]{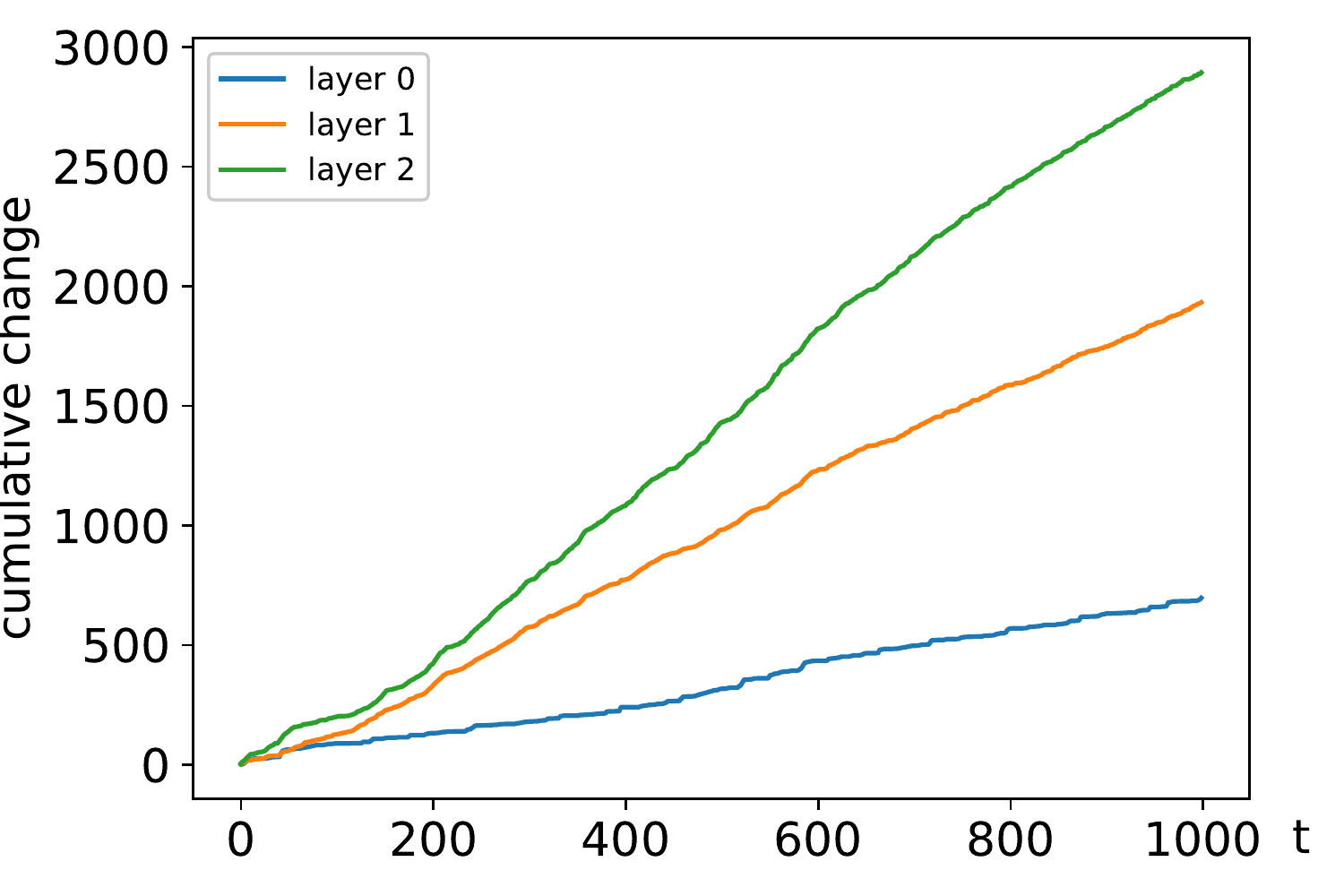}
  \end{subfigure}
  \hfill
  \begin{subfigure}[t]{0.325\textwidth}
      \centering
      \includegraphics[width=\textwidth]{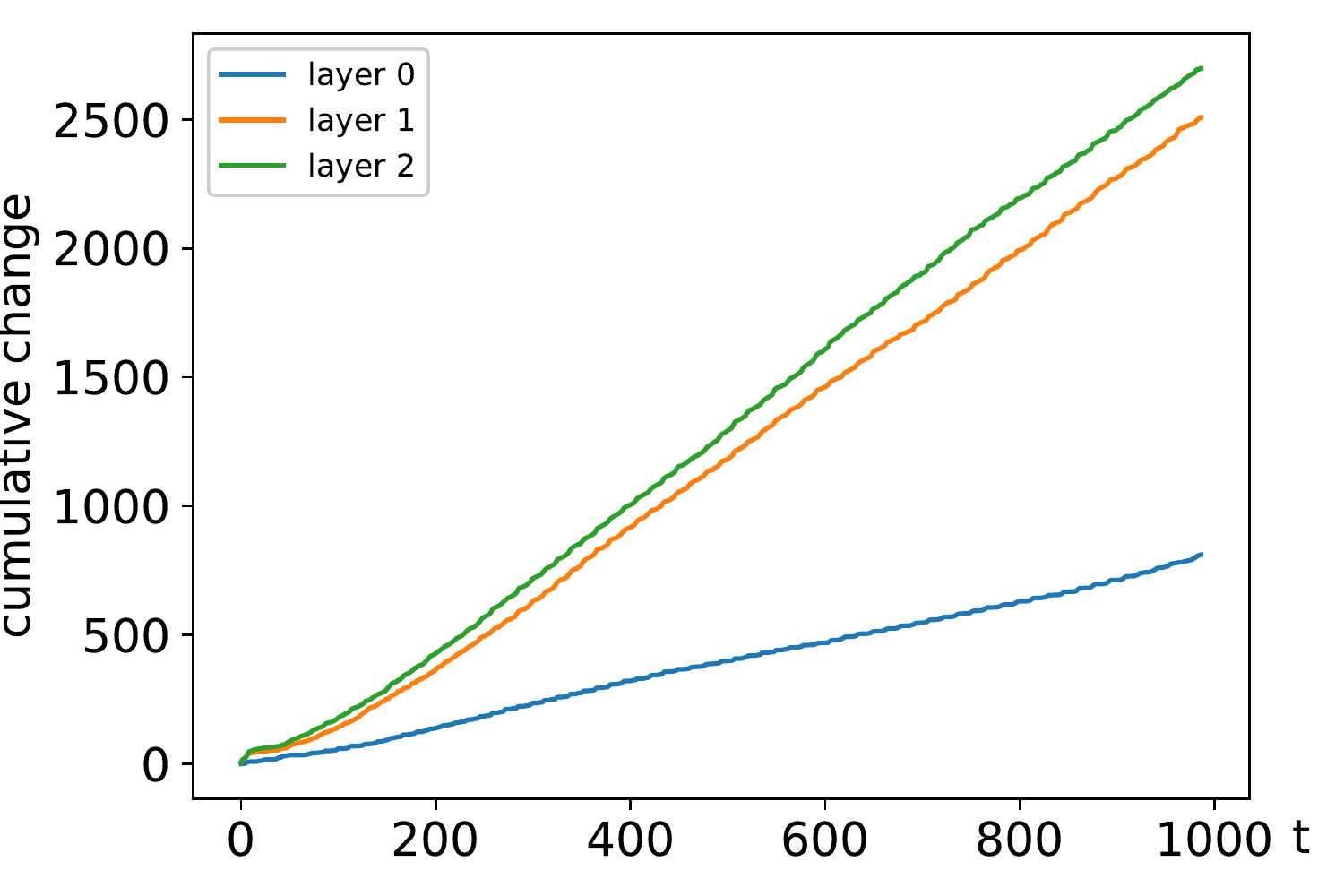}
  \end{subfigure}
  \hfill
  \begin{subfigure}[t]{0.325\textwidth}
      \centering
      \includegraphics[width=\textwidth]{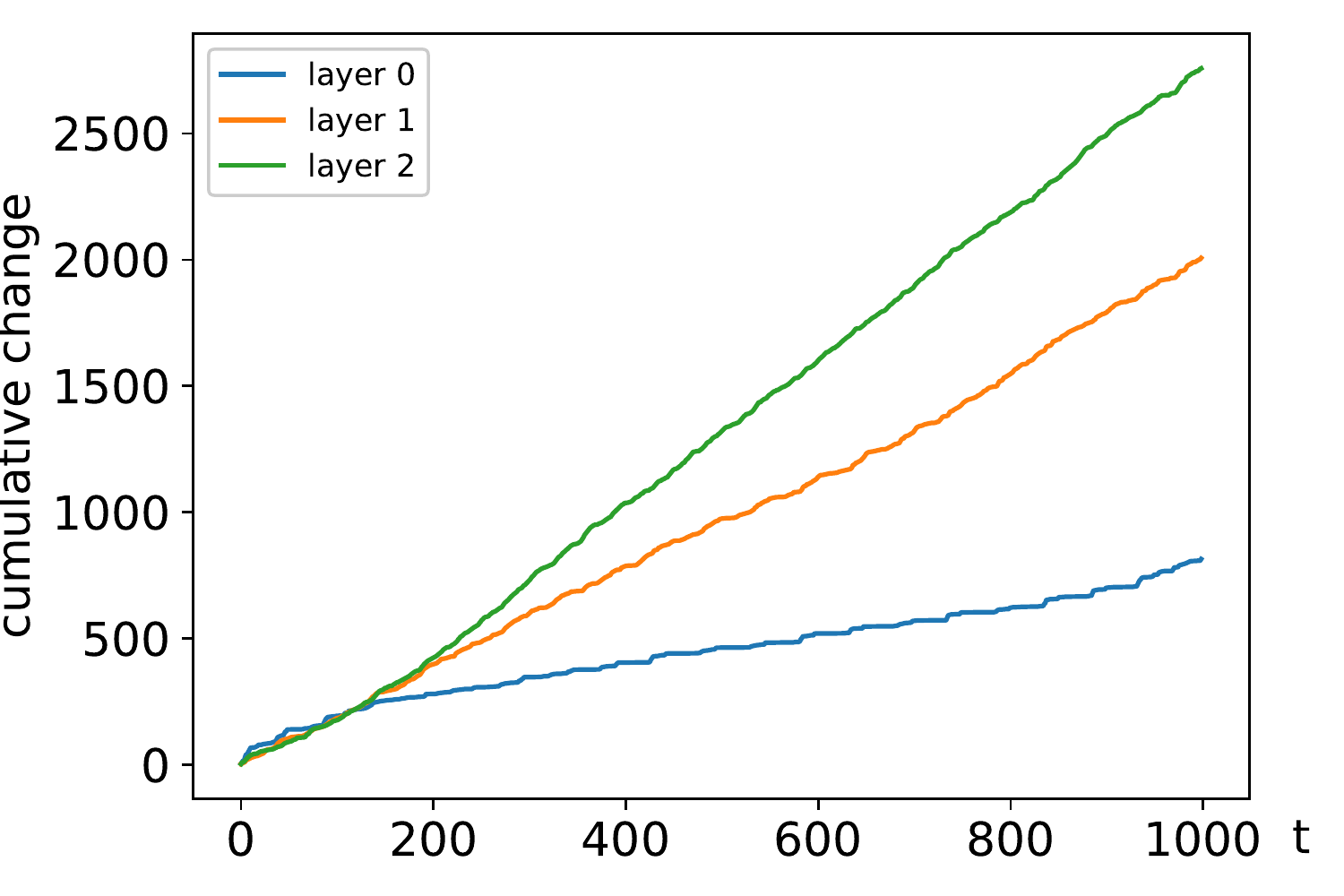}
  \end{subfigure}
  \caption{Absolutive cumulative change in the attention masks for three different models on \texttt{Walker-7}.}
  \label{fig:walker-cumulative-change}
\end{figure}

\section{Generalisation results}
\label{sec:generalisation-results}

\begin{table}[h]
  \centering
  \caption{Initial results on generalisation. The numbers show the average performance of three seeds evaluated on 100 rollouts and standard error of the mean. While the average values are higher for \method{} on 5 out of 7 benchmarks, high variance of both methods might be indicative of instabilities in generalisation behaviour due to large differences between the training and testing tasks.}
  \label{tab:generalisation-results}
  \begin{tabular}{lll}
  \hline
  \textbf{}                                    & \multicolumn{1}{l}{\meth{amorpheus}} & \multicolumn{1}{l}{\meth{smp}} \\ \hline
  \texttt{walker}-\texttt{3}-\texttt{main}                                & \textbf{666.24} (133.66)                        & 175.65 (157.38)                  \\
  \texttt{walker}-\texttt{6}-\texttt{main}                                & \textbf{1171.35} (832.91)                       & 729.26 (135.60)                  \\
  \midrule
  \texttt{humanoid}-\texttt{2d}-\texttt{7}-\texttt{left}-\texttt{leg}                       & \textbf{2821.22} (1340.29)                      & 2158.29 (785.33)                 \\
  \multicolumn{1}{r}{\texttt{humanoid}-\texttt{2d}-\texttt{8}-\texttt{right}-\texttt{knee}} & \textbf{2717.21} (624.80		)                      & 327.93 (125.75)                  \\
  \midrule
  \texttt{cheetah}-\texttt{3}-\texttt{balanced}                          &  \textbf{474.82}	(74.05)     & 156.16 (33.00)                                  \\
  \texttt{cheetah}-\texttt{5}-\texttt{back}                               & \multicolumn{1}{l}{3417.72		 (306.84)}                   & \multicolumn{1}{l}{\textbf{3820.77} (301.95)}             \\
  \texttt{cheetah}-\texttt{6}-\texttt{front}                              &   5081.71 (391.08) &          \textbf{6019.07} (506.55)                        \\ \hline
  \end{tabular}%
  \end{table}

\section{Residual Connection Ablation}

\begin{figure}[h]
  \centering
  \begin{subfigure}[t]{0.32\textwidth}
      \centering
      \includegraphics[width=\textwidth]{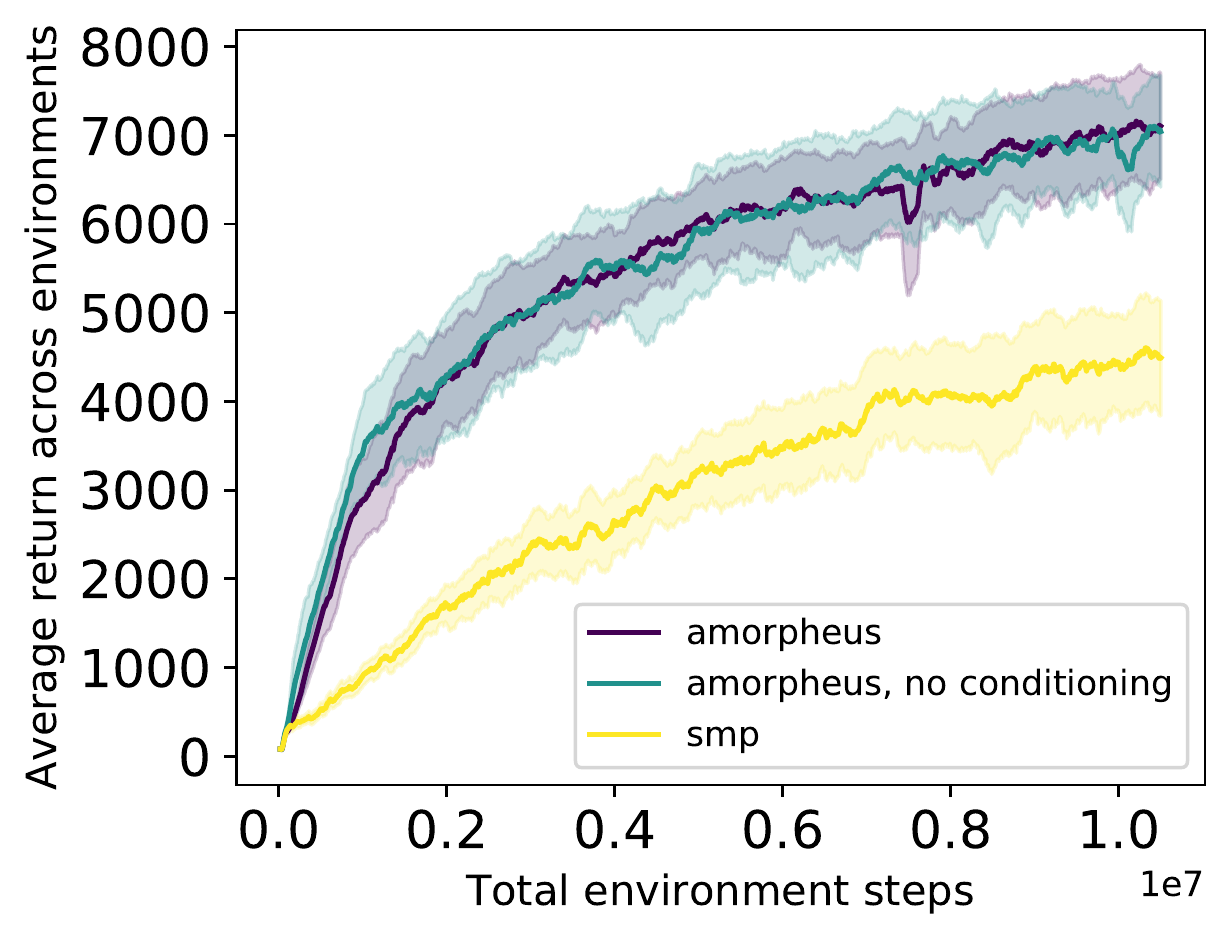}
      \caption{Walker++}
  \end{subfigure}
  \hfill
  \begin{subfigure}[t]{0.32\textwidth}
      \centering
      \includegraphics[width=\textwidth]{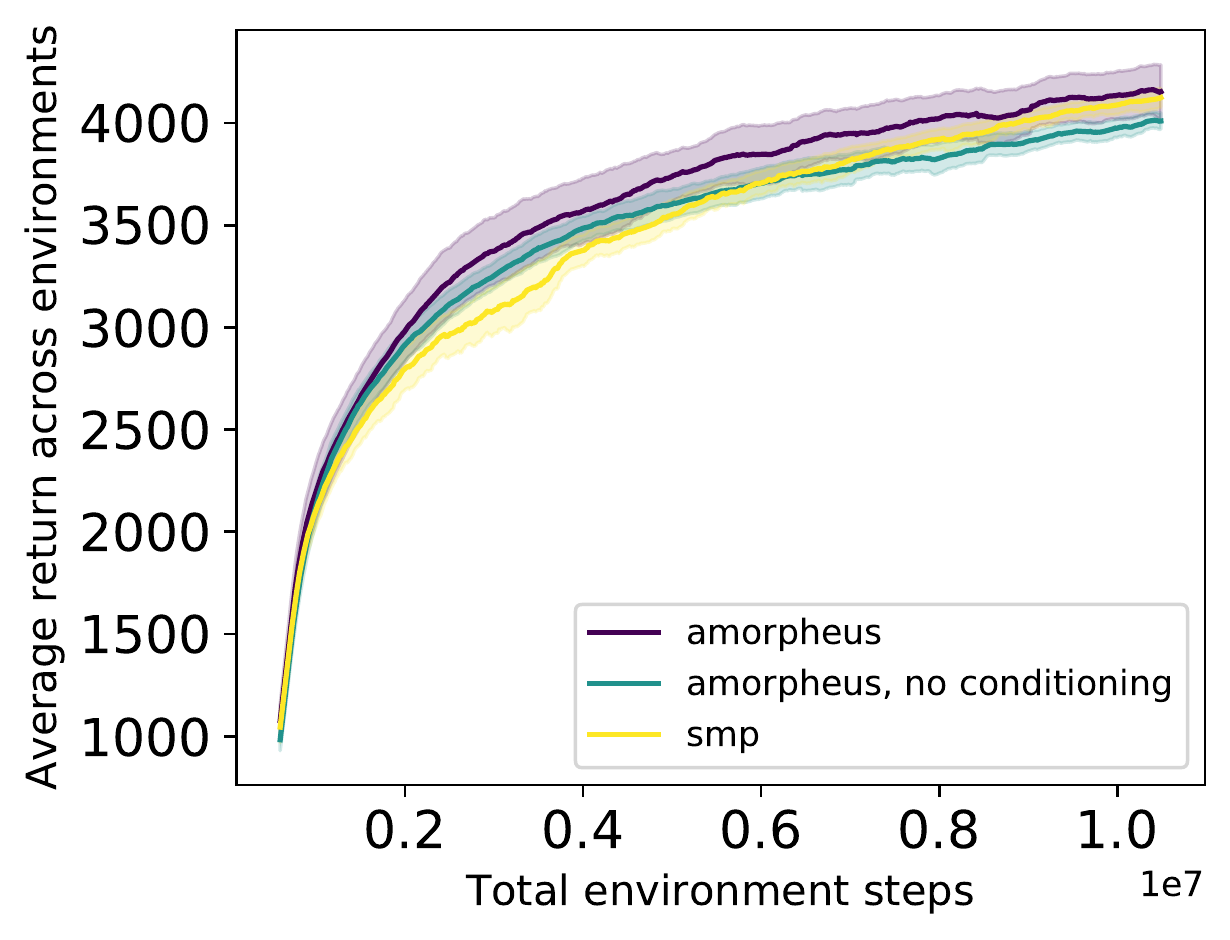}
      \caption{Cheetah++}
  \end{subfigure}
  \hfill
  \begin{subfigure}[t]{0.32\textwidth}
      \centering
      \includegraphics[width=\textwidth]{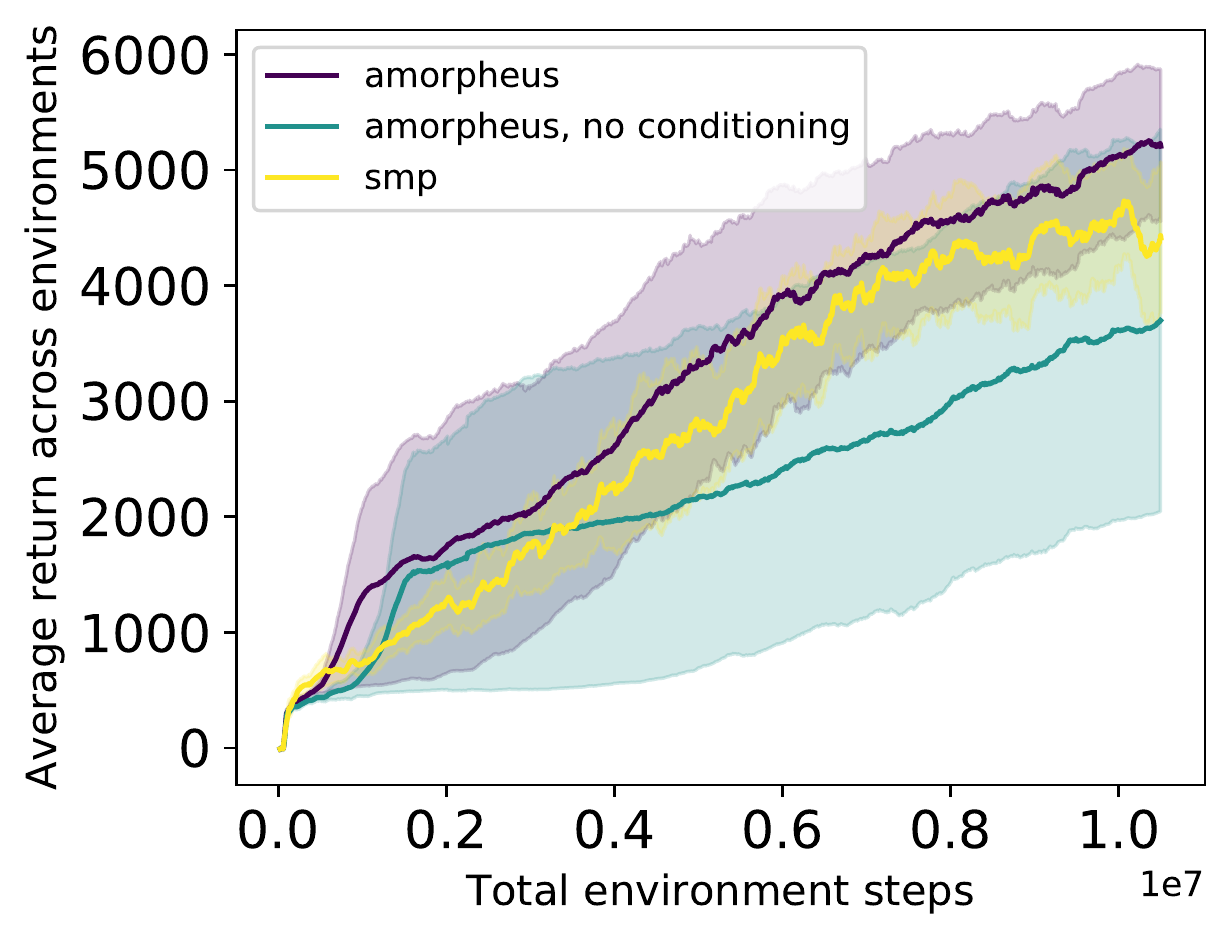}
      \caption{Humanoid++}
  \end{subfigure}
  \hfill
  \begin{subfigure}[t]{0.32\textwidth}
      \centering
      \includegraphics[width=\textwidth]{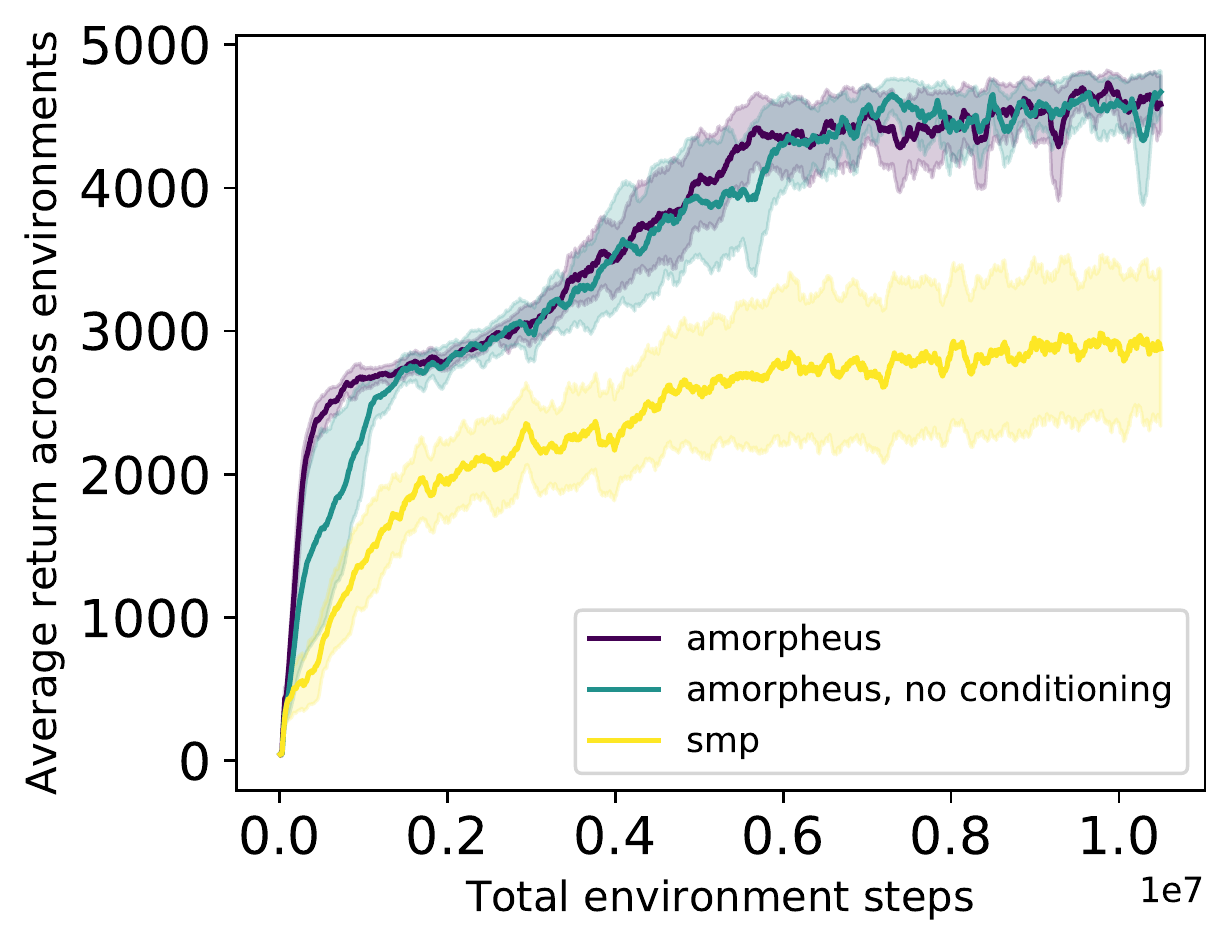}
      \caption{Hopper++}
  \end{subfigure}
  \begin{subfigure}[t]{0.32\textwidth}
    \centering
    \includegraphics[width=\textwidth]{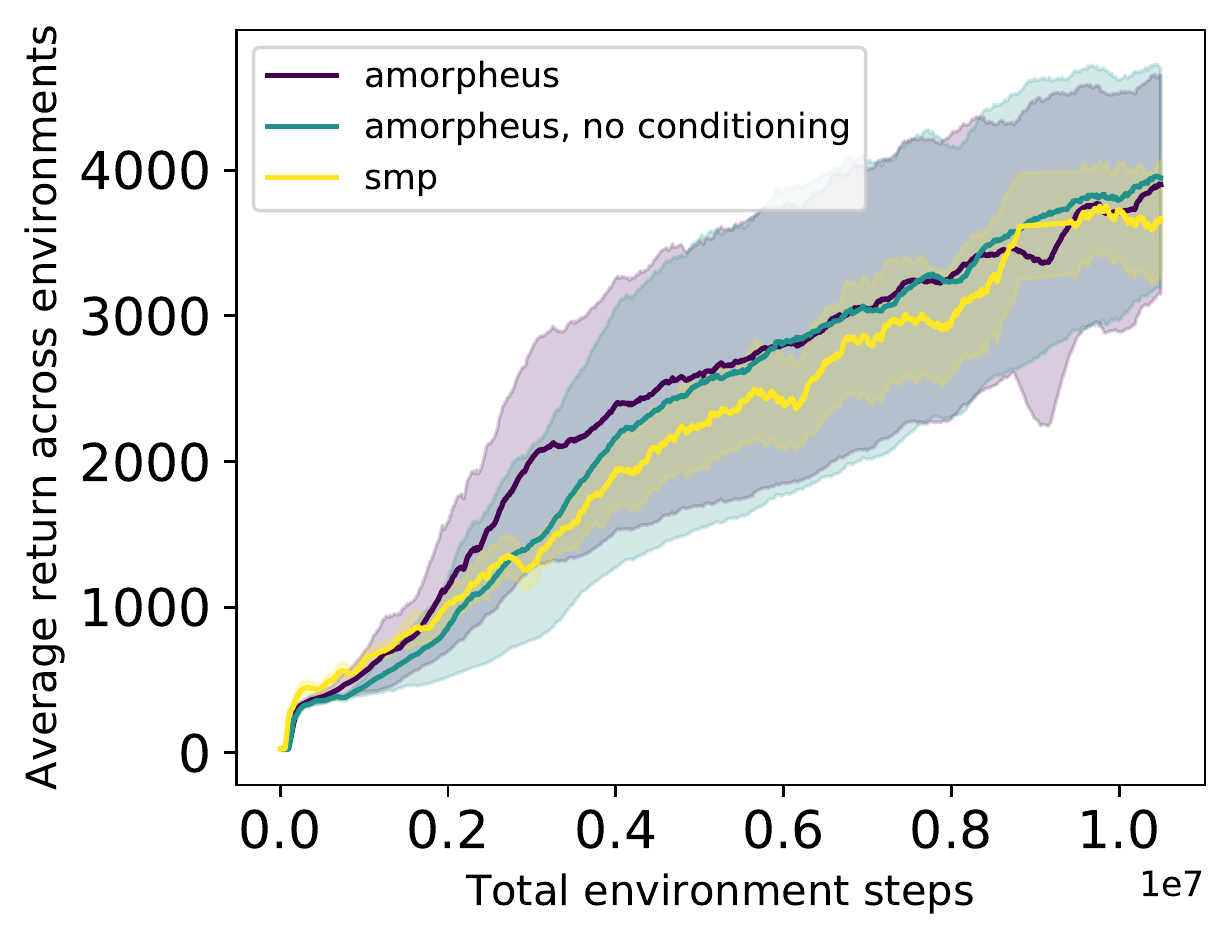}
    \caption{\scriptsize Walker-Humanoid++}
  \end{subfigure}
  \begin{subfigure}[t]{0.32\textwidth}
    \centering
    \includegraphics[width=\textwidth]{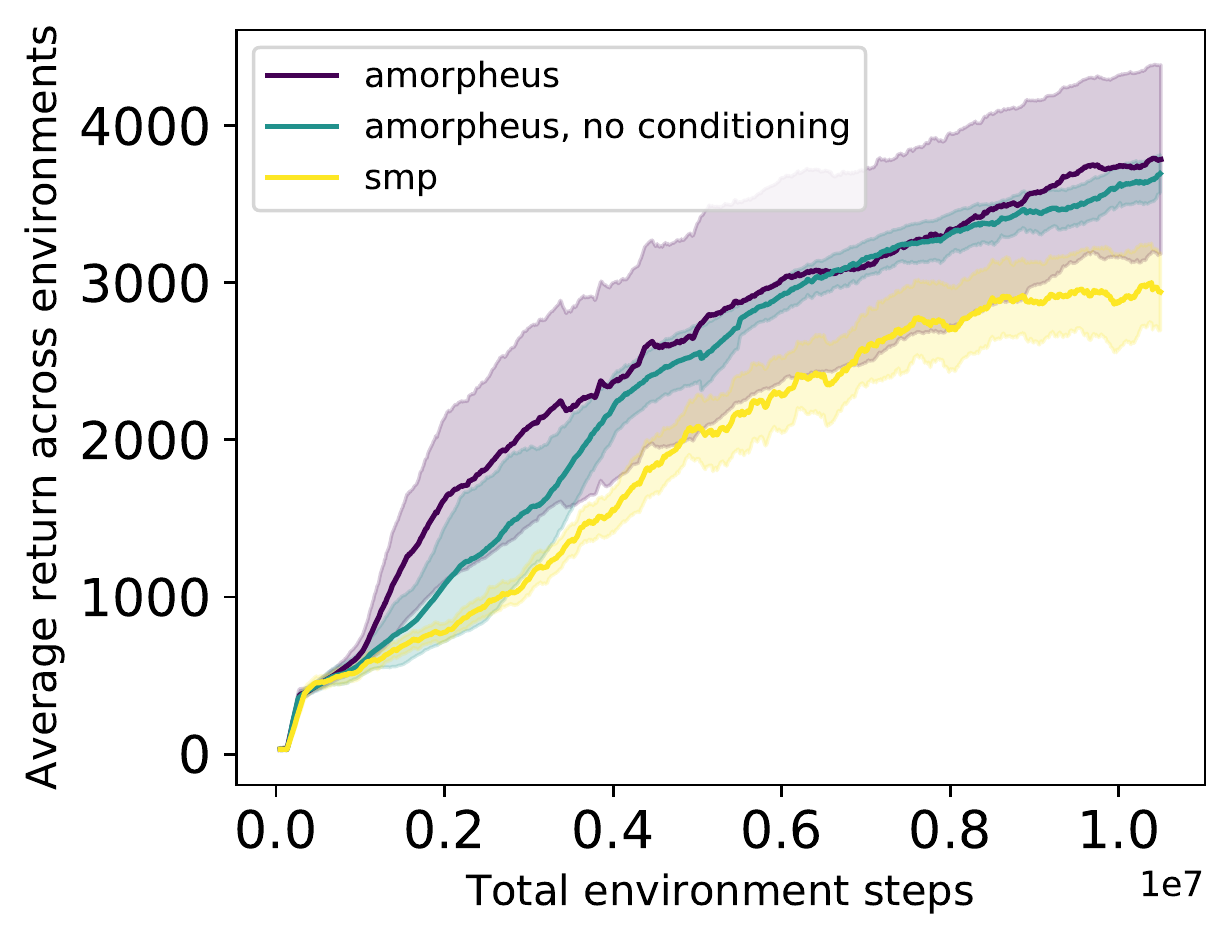}
    \caption{\scriptsize Walker-Humanoid-Hopper++}
  \end{subfigure}
  \begin{subfigure}[t]{0.42\textwidth}
      \centering
      \includegraphics[width=\textwidth]{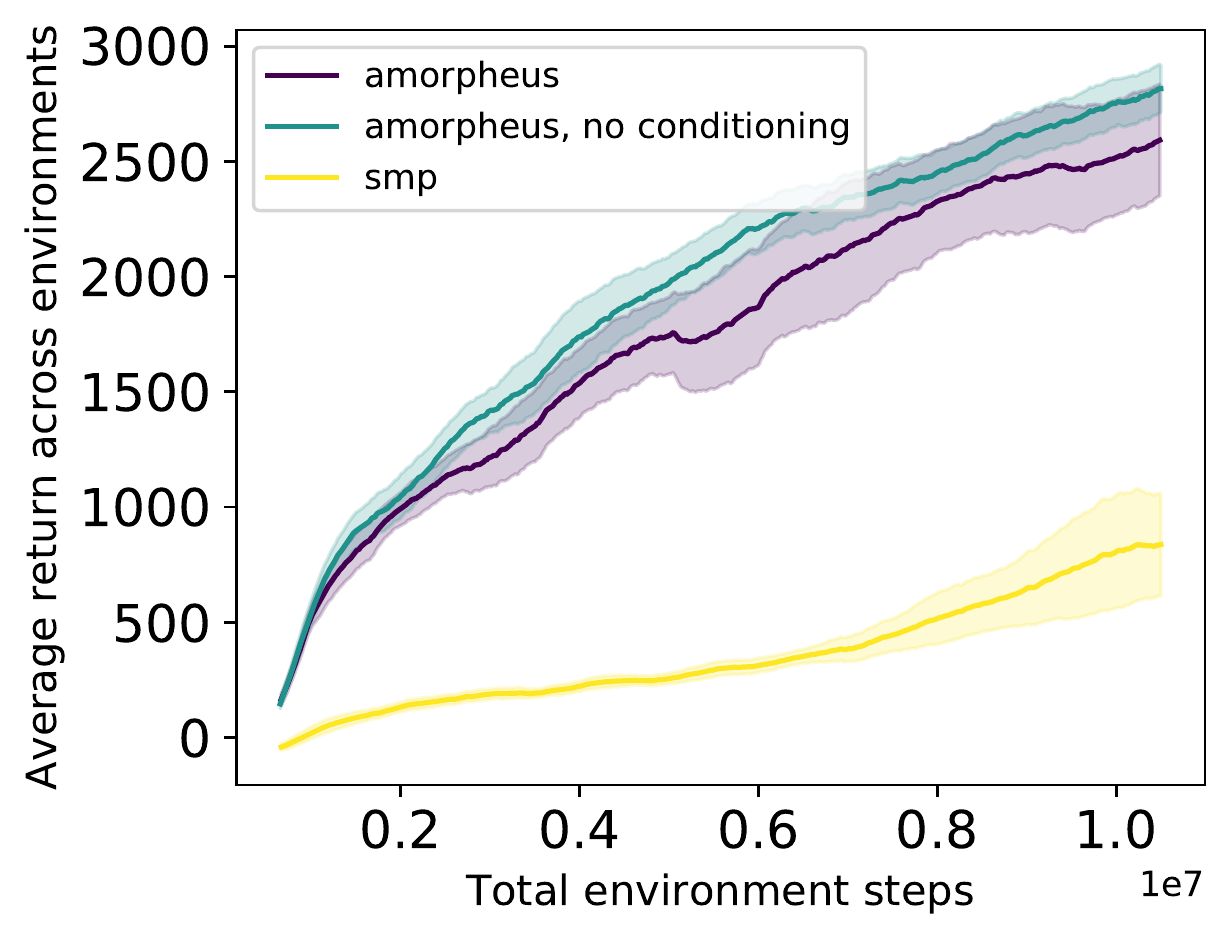}
      \caption{Cheetah-Walker-Humanoid++}
  \end{subfigure}
  \begin{subfigure}[t]{0.42\textwidth}
      \centering
      \includegraphics[width=\textwidth]{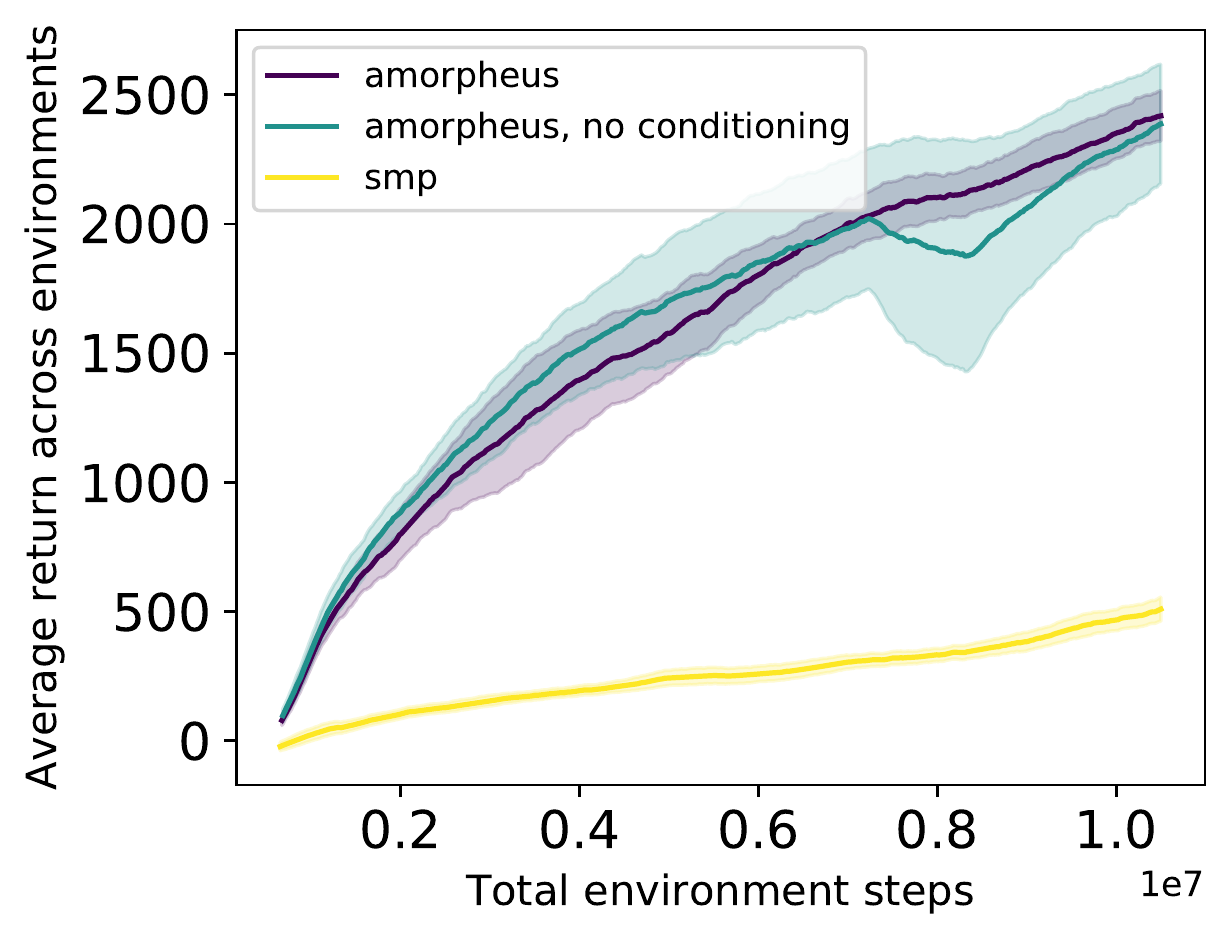}
      \caption{\scriptsize Cheetah-Walker-Humanoid-Hopper++}
  \end{subfigure}
  \caption{Residual connection ablation experiment.}
  \label{fig:residual-ablation}
\end{figure}

We use the residual connection in \method{} as a safety mechanim to prevent nodes from forgetting their own observations.
To check that \method{}'s improvements do not come from the residual connection alone, we performed the ablation.
As one can see on Figure~\ref{fig:residual-ablation}, we cannot attribute the success of our method to this improvement alone.
High variance on \texttt{Humanoid++} is related to the fact that one seed started to improve much later, and the average performance suffered as the result.

\end{document}